\pdfoutput=1

\documentclass[11pt]{article}

\usepackage[table]{xcolor}
\usepackage{acl}  

\usepackage{times}
\usepackage{latexsym}

\usepackage[T1]{fontenc}

\usepackage[utf8]{inputenc}

\usepackage{microtype}

\usepackage{inconsolata}

\usepackage{graphicx}

\usepackage{tabularx}
\usepackage{amsmath}
\usepackage{booktabs}
\usepackage{multirow}
\usepackage{geometry}
\usepackage{siunitx}
\usepackage[table]{xcolor}
\usepackage{dashrule}
\usepackage{verbatim}

\usepackage{tcolorbox}
\usepackage{fancyvrb}
\usepackage{fvextra}
\tcbuselibrary{breakable}
\usepackage{paralist}

\usepackage{hyperref}
\usepackage{svg} 

\usepackage{amsmath, amssymb, amsthm}
\theoremstyle{plain}
\newtheorem{theorem}{Theorem}
\newtheorem{theorem_appendix}{Theorem}

\title{DivScore: Zero-Shot Detection of LLM-Generated Text \\ in Specialized Domains}

\author{
 \textbf{Zhihui Chen\textsuperscript{1}\thanks{Equal contribution}},
 \textbf{Kai He\textsuperscript{1}\footnotemark[1]},
 \textbf{Yucheng Huang\textsuperscript{2}},
 \textbf{Yunxiao Zhu\textsuperscript{3}},
 \textbf{Mengling Feng\textsuperscript{1}\thanks{Corresponding author}}
\\
\\
 \textsuperscript{1}Saw Swee Hock School of Public Health, National University of Singapore\\
 \textsuperscript{2}Interactive Entertainment Group, Tencent,\\
 \textsuperscript{3}Faculty of Law, The University of Hong Kong  \\
\texttt{zhihui.chen@u.nus.edu},
\texttt{huangyucheng.andrew@gmail.com}, \\
\texttt{talia\_zhu@connect.hku.hk}, \texttt{\{kai\_he,ephfm\}@nus.edu.sg}  
}

\begin{document}
\maketitle
\begin{abstract}
Detecting LLM-generated text in specialized and high-stakes domains like medicine and law is crucial for combating misinformation and ensuring authenticity. However, current zero-shot detectors, while effective on general text, often fail when applied to specialized content due to domain shift. We provide a theoretical analysis showing this failure is fundamentally linked to the KL divergence between human, detector, and source text distributions. To address this, we propose DivScore, a zero-shot detection framework using normalized entropy-based scoring and domain knowledge distillation to robustly identify LLM-generated text in specialized domains. 
We also release a domain-specific benchmark for LLM-generated text detection in the medical and legal domains.
Experiments on our benchmark show that DivScore consistently outperforms state-of-the-art detectors, with 14.4\% higher AUROC and 64.0\% higher recall (0.1\% false positive rate threshold). In adversarial settings, DivScore demonstrates superior robustness than other baselines, achieving on average 22.8\% advantage in AUROC and 29.5\% in recall. Code and data are publicly available\footnote{https://github.com/richardChenzhihui/DivScore}.
\end{abstract}

\section{Introduction}

As Large Language Models (LLMs) produce text of increasing fluency and realism, the need for reliable machine-generated text detection has become critical. This is especially true in high-stakes domains like healthcare and law, where undetected AI-written content could undermine trust and safety. 

Initial approaches to detecting LLM-generated text employed supervised learning, such as GPT-2 detector~\cite{loglik-roberta-solaiman2019release} and GPT-3.5 detector~\cite{hc3}.
While effective on the data they were trained on, such supervised detectors show clear shortcomings when faced with new domains or more advanced language models. 
For instance, a detector trained on law articles may falter on healthcare text, and a classifier tuned to GPT-2 outputs may misclassify text from a newer LLM. 
Furthermore, every new LLM or domain requires the creation of a fresh labeled dataset and retraining of the detector, rendering the process resource-intensive and inefficient.

In response, researchers have developed detectors that require no training examples and instead exploit statistical irregularities of LLM-generated text. For instance, DetectGPT~\cite{detectgpt} uses an LLM’s probability curvature to flag generated passages without any finetuning, while Binoculars~\cite{Binoculars} contrasts the outputs of two language models to achieve state-of-the-art detection accuracy with zero training data.


\begin{figure}
    \centering
    \includegraphics[width=1\linewidth]{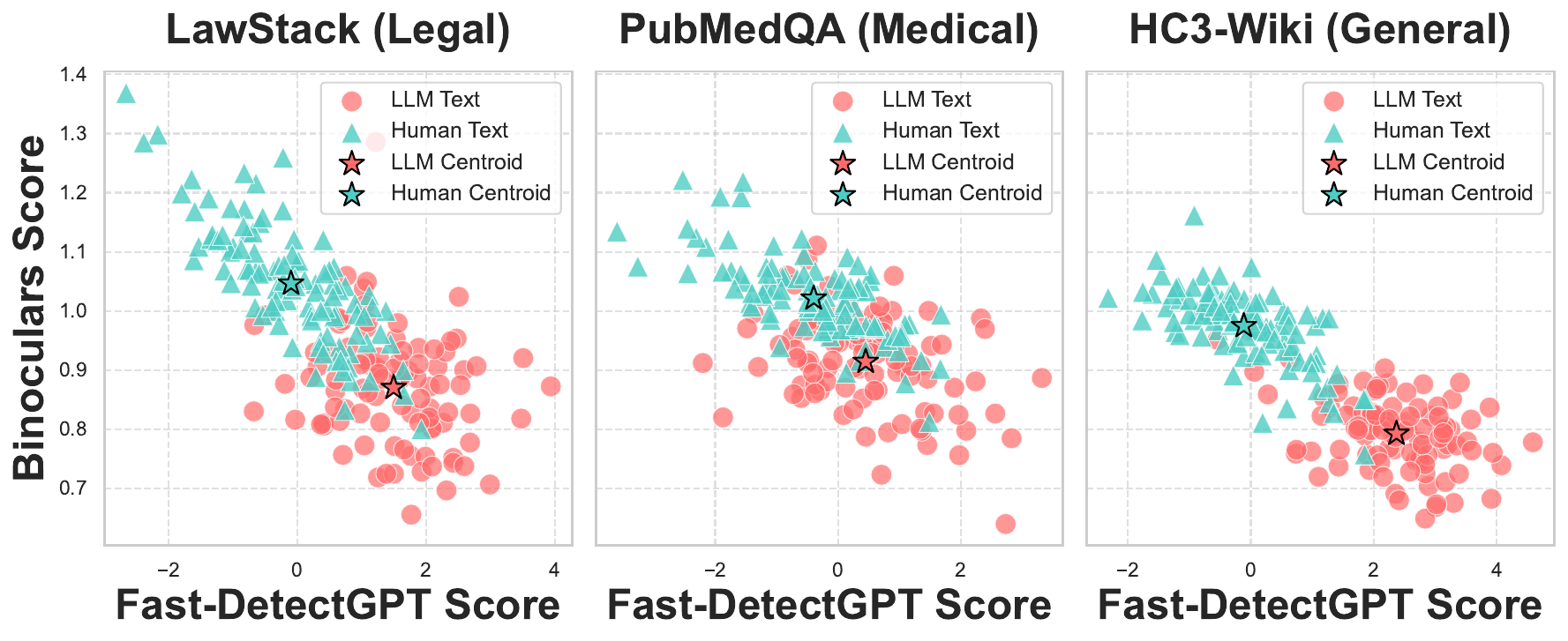}
    \caption{Fast-DetectGPT~\cite{fastdetectgpt} and Binoculars~\cite{Binoculars} detector scores of LLM/Human texts in specialized (legal, medical) and general domain. In specialized domain, both detectors failed in creating scoring gap necessary for detecting LLM-generated text}
    \label{fig: special_signature}
\end{figure}

However, most existing zero-shot detection studies focus on general domain data. 
Specialized domains such as healthcare literature or legal documents present several additional challenges that make LLM-generated text detection more difficult. 
These domains typically have: (1) Sparse labeled data. Few or no annotated examples of LLM-generated text for training. (2) Complex language. Domain-specific jargon and intricate writing styles that differ markedly from common text. (3) Unfamiliar content. Topic areas and facts that even advanced LLMs may not have seen frequently during pre-training.

As a result, detection methods that perform well on general text often struggle under such domain shift. As shown in Figure 1, we observe that Fast-DetectGPT~\cite{fastdetectgpt} and Binoculars~\cite{Binoculars} method both tend to compress in specialized domains: both human and LLM-generated texts yield similar detector scores, eroding the gap needed for reliable discrimination. To address these gaps, we propose DivScore – a normalized entropy-based divergence metric for domain-specific, zero-shot detection of LLM-generated text in specialized domains. 
Our approach is specifically designed to be robust across different domains.  

In essence, DivScore works by normalizing the domain-specific entropy of a candidate text against an estimated baseline, thereby measuring how surprising the text is relative to what a reference language model expects for that domain. 
Our method first highlights domain-specific alignment, and then applies entropy normalization to accentuate the differences between human-written and LLM-generated text, thereby enhancing the contrast necessary for detection.
We implement DivScore using an unsupervised domain knowledge distillation framework, maintaining a zero-shot setting that requires no labeled data.
In our evaluations, DivScore outperforms state-of-the-art (SOTA) zero-shot baselines under significant domain shifts, while remaining competitive in general domain settings.
We make the following key contributions:
\begin{itemize}
\item We introduce \textbf{DivScore}. To the best of our knowledge, it is the first zero-shot detection method designed for identifying domain-specific LLM-generated text, inspired by domain alignment and entropy normalization.
\item We release a high-quality benchmark for detecting LLM-generated text in specialized domains. It includes professionally legal and medical corpora, along with knowledge distillation datasets and an adaptation pipeline for detector evaluation and finetuning.  
\item We conduct extensive experiments and analysis. The results show that DivScore outperforms SOTA baselines by 14.4\% in AUROC and 64.0\% in recall (0.1\% false positive rate), while also exhibiting strong robustness under adversarial conditions.
\end{itemize}

\section{Related Work}
Machine-generated text detection is typically formulated as a binary classification problem~\cite{detectllm,fastdetectgpt,detectgpt,loglik-roberta-solaiman2019release,Binoculars}. 
Supervised detectors, such as RoBERTa-based models~\cite{loglik-roberta-solaiman2019release}, perform well on in-domain data, but require additional manual annotation, with limited scalability and flexibility~\cite{supervised_ood,detectllm,detectgpt}. 

Another group of studies aims to construct zero-shot detectors based on the discrepancy in LLM-measured statistical signatures between human and machine-generated text. 
Text statistics-based detectors such as LLM text entropy~\cite{entropy-lavergne2008detecting}, log-probability~\cite{loglik-roberta-solaiman2019release} and per-token rank~\cite{rank_logrank-gehrmann2019gltr} have proven to be effective in general LLM text detection. 
\citet{detectgpt} proposed DetectGPT to detect LLM-generated passages by the curvature of LLM text log-probability function under controlled text perturbations. 
Fast-DetectGPT~\cite{fastdetectgpt} substitutes the perturbation method in DetectGPT with a more efficient LLM sampling step. 
\citet{Binoculars} further discovered that contrasting the LLM text perplexity of two closely related language models can effectively distinguish LLM-generated content. 

However, the zero-shot property comes at the cost of limited adaptability in specialized domains. Current zero-shot detectors,  including Binoculars~\cite{Binoculars}, mainly focus on general domain content, such as news, essays, and social media posts~\cite{xsum-BBC, hc3, writeprompt_story, LLM_survey, RAID_benchmark}. Relying on general domain LLMs for statistical signature scoring, current methods implicitly assume sufficient domain familiarity from the detector. In specialized domains with rare knowledge (e.g., medical or legal), both human and LLM-generated texts could be unfamiliar to these detectors, collapsing the statistical gap and degrading detection performance. Our DivScore is designed for avoiding such problems.

\begin{figure}[t]
    \centering
    \includegraphics[width=\linewidth]{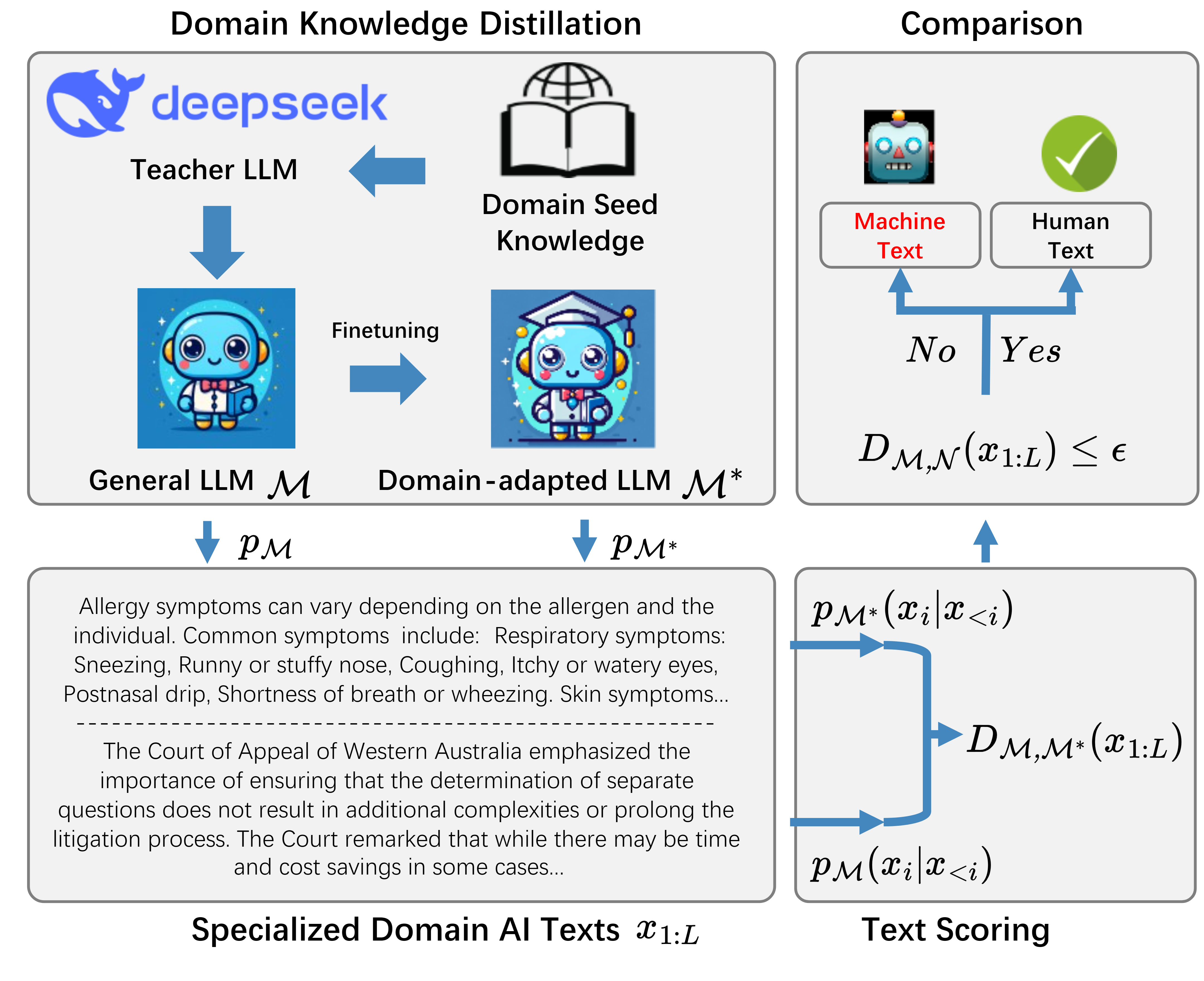}
    \caption{The overall framework of DivScore}
    \label{fig: main}
\end{figure}

\section{Methodology}
\label{section:method}

In this section, we first formalize why zero-shot detectors falter when the text distribution shifts to a specialised domain, setting the theoretical stage for \textbf{DivScore}. Guided by this analysis, we derive our core detector and then introduce a domain-knowledge distillation step that narrows the gap between the detector and specialised language, markedly strengthening robustness and accuracy. The framework of DivScore is shown in Figure~\ref{fig: main}.

\subsection{Theoretical Analysis: Why Do Zero-Shot Detectors Fail under Domain Shift?}

Most zero-shot detectors, such as entropy-based scoring~\cite{entropy-lavergne2008detecting} or perplexity-based Binoculars~\cite{Binoculars}, assume that LLM-generated text exhibits statistical signatures (e.g., lower entropy, distinctive perplexity) that are distinguishable from human-written text. 

However, when applied to corpus from specialized domains (see Figure~\ref{fig: special_signature}), these signals often collapse: both human- and machine-generated texts may appear equally "surprising" or "familiar" to a general domain LLM. To formalize this phenomenon, we firstly analyse the theoretical performance of a general zero-shot detector that scores a text $X$ based on its likelihood $p_Q(X)$ with a detector LLM. 
Specifically, we present the following theorem, which characterizes the relationship between detector performance and the two specific Kullback–Leibler (KL) divergence.

\begin{theorem}

\label{theorem:1}
Given a zero-shot LLM-generated text detector $D$ with detection score mapped from the detector LLM's text likelihood $p_{Q}(X)$ by any strictly monotonic function $f: \mathbb{R} \to \mathbb{R}$ :
\begin{equation}
D(X) = f(p_{Q}(X)),
\end{equation}
where $Q$ represents the text distribution calculated from $D$.
The area under the receiver operating characteristic curve (AUROC) of $D$, denoted as $\textup{AUROC}(D(X))$, satisfies:
\begin{equation}
\textup{AUROC}(D(X)) \propto^{+} \delta_{KL},
\end{equation}
where $\propto^+$ denotes positive monotonic association, and $\delta_{KL}$, effective KL divergence, is given as: 
\begin{equation}
\delta_{KL} =  D_{\text{KL}}(P \parallel Q') - D_{\text{KL}}(Q \parallel Q'),
\end{equation}
with $Q'$ indicating text distribution from the real source. $P$ denotes text distribution from human. $D_{KL}(\cdot||\cdot)$ denotes KL divergence.
\end{theorem}

Theorem~\ref{theorem:1} shows that the performance of a zero-shot detector depends on the extent to which the human text distribution $P$ diverges more from the real source distribution $Q'$ than from the detector distribution $Q$. 
In specialized domains (e.g., medical or legal) with distinctive linguistic characteristics, the source distribution $Q'$ may more closely resemble $P$ than the general-domain LLM distribution $Q$. 
Consequently, a reduction in $\delta_{KL}$ diminishes the detector's ability to distinguish machine-generated text within such domains.
See detailed theorem proof in Appendix~\ref{section: theorem_proof}.

\subsection{DivScore: Normalized Entropy-based Scoring}
\label{section: divscore}
Building on the theoretical analysis, we design a detector that leverages the entropy of a domain-adapted LLM $\mathcal{M}^*$ (see subsection~\ref{section: method_distill}) to quantify how predictable or familiar a sentence $x_{1:L}$ is under its distribution. Given a $x_{1:L} = (x_1, ..., x_L)$, the mean token entropy is defined as:
\begin{equation}
\label{equation: entropy}
\begin{split}
H_{\mathcal{M}^*}(x_{1:L}) 
&= -\frac{1}{L}
   \sum_{i=1}^{L} p_{\mathcal{M^*}}(x_i \mid x_{<i})\\
&\quad\;\cdot \log p_{\mathcal{M}^*}(x_i \mid x_{<i})
\end{split}
\end{equation}
Ideally, if $\mathcal{M}^*$ is well aligned with the distribution of LLM-generated text in the designated domain, LLM-generated text should yield lower entropy than human-written text. However, in practice, this assumption may not always hold due to subtle domain mismatches or the presence of highly predictable (i.e., low-entropy) human texts.

To address this, we introduce a normalization scheme based on the cross-entropy between a general LLM $\mathcal{M}$ and the domain-adapted LLM $\mathcal{M}^*$. The cross-entropy is defined as:
\begin{equation}
\label{equation: ce}
\begin{split}
CE_{\mathcal{M},\mathcal{M}^*}(x_{1:L})
&= -\frac{1}{L}
   \sum_{i=1}^{L} p_{\mathcal{M}}(x_i \mid x_{<i})\\
&\quad\;\cdot \log p_{\mathcal{M}^*}(x_i \mid x_{<i})
\end{split}
\end{equation}
The above cross-entropy quantifies the degree of disagreement between the two models' probability estimates for text. For LLM-generated text in a specialized domain, the domain-adapted $\mathcal{M}^*$ is expected to diverge significantly from the general LLM $\mathcal{M}$, resulting in a high cross-entropy score. In contrast, for human-written text that is unfamiliar to both models, the disagreement is typically lower, leading to a lower cross-entropy value. This intuition is further supported by a detailed statistical analysis in Section~\ref{section: analysis_distill}.
The final DivScore is computed as:
\begin{equation}
\label{equation: detector}
    D_{\mathcal{M},\mathcal{M}^*}(x_{1:L}) = \frac{H_{\mathcal{M}^*}(x_{1:L})}{CE_{\mathcal{M},\mathcal{M}^*}(x_{1:L})}
\end{equation}

A higher $D_{\mathcal{M},\mathcal{M}^*}(x_{1:L})$ driven by high entropy (unfamiliarity) and low cross-entropy (low model disagreement) - indicates that $x_{1:L}$ is more likely to be human-written. Conversely, a lower score suggests that the text is machine-generated. This DivScore normalization enhances the separation between human and LLM-generated text, especially in specialized domains where raw entropy-based signals alone are insufficient.

\subsection{Enhancing the Detector via Unsupervised Domain Knowledge Distillation}
\label{section: method_distill}
DivScore employs a paired LLM $(\mathcal{M}, \mathcal{M}^*)$, consisting of a general-purpose and domain-adapted model, for scoring.
In the basic DivScore setup, $\mathcal{M}^*$ can be any LLM adapted for the target domain. 
Further, motivated by Theorem~\ref{theorem:1} and aiming to enhance detection performance in highly specialized domains, we introduce a tailored improvement: constructing $\mathcal{M}^*$ by distilling domain-specific knowledge into the general model $\mathcal{M}$.
The goal of this enhancement is to reduce the KL divergence $D_{\text{KL}}(Q \parallel Q')$ by aligning the detector's distribution with the domain-specific LLM-generated text. The process is as follows:

\begin{enumerate}
    \item \textbf{Teacher Model Selection:} Select a large language model (LLM) with extensive domain expertise and a substantial parameter count to act as the teacher model (e.g., DeepSeek-R1).
    \item \textbf{Domain Seed Knowledge Construction:} Collect a set of questions representative of the specialized domain. Using seed knowledge, prompt the teacher LLM to generate detailed and in-depth questions based on the initial input.
    \item \textbf{Distillation Knowledge Generation:} Use the teacher LLM to generate high-quality answers and narrative explanation to the seed questions, forming a domain-specific distillation knowledge base consisting of outputs from the teacher LLM.
    \item \textbf{Student Learning:} Finetune the general LLM $\mathcal{M}$ of DivScore with the generated domain-specific question-answer pairs, yielding the enhanced model $\mathcal{M}^*$ adapted to this specialized domain.
\end{enumerate}

This distillation process for building $\cal{M^*}$ transfers domain knowledge from a teacher LLM to the general LLM $\cal{M}$ of DivScore, narrows the gap between the detector and the LLM text distributions, and thus increases $\delta_{KL}$ in Theorem~\ref{theorem:1}, improving  detection performance. 
The entire approach operates in a zero-shot manner, requiring no human-labeled text for training and inference, which makes it highly practical for real-world deployment in specialized domains.

We release all the above resources as a benchmark, which consists of (1) domain-specific knowledge distillation datasets and adaptation pipelines for medical and legal domains, and (2) 4,000 high-quality professional legal and medical documents paired with LLM-generated counterparts for detector evaluation.

\begin{table*}[ht]
\centering
\resizebox{\linewidth}{!}{%
\large
\setlength{\tabcolsep}{4.5pt}
\renewcommand{\arraystretch}{1.15}
\begin{tabular}{l 
|cccc|cccc|cccc|cccc}
\toprule
\textit{Text Source}
& \multicolumn{4}{c}{GPT-4o} 
& \multicolumn{4}{c}{O3-mini} 
& \multicolumn{4}{c}{DeepSeek-V3} 
& \multicolumn{4}{c}{DeepSeek-R1} \\
\cmidrule(lr){2-5} \cmidrule(lr){6-9} \cmidrule(lr){10-13} \cmidrule(lr){14-17} 
\textit{Datasets} & MIMIC & PubMed & OALC & LawStack 
& MIMIC & PubMed & OALC & LawStack
& MIMIC & PubMed & OALC & LawStack
& MIMIC & PubMed & OALC & LawStack  \\
\midrule
\multicolumn{1}{c}{}  & \multicolumn{16}{c}{AUROC} \\
\midrule
RoB-base        & $66.79_{(\downarrow31.55)}$ & $85.79_{(\downarrow9.63)}$ & $58.33_{(\uparrow19.86)}$ & $65.10_{(\downarrow29.98)}$ & $67.98_{(\downarrow26.71)}$ & $98.70_{(\uparrow14.13)}$ & $58.07_{(\uparrow0.67)}$ & $73.35_{(\downarrow9.02)}$ & $69.11_{(\downarrow30.89)}$ & $82.39_{(\downarrow17.33)}$ & $23.20_{(\downarrow75.07)}$ & $56.54_{(\downarrow43.33)}$ & $68.99_{(\downarrow20.11)}$ & $99.63_{(\uparrow9.46)}$ & $65.24_{(\uparrow10.69)}$ & $74.73_{(\downarrow13.16)}$  \\
RoB-large      & $93.05_{(\downarrow5.29)}$ & $91.69_{(\downarrow3.73)}$ & $81.23_{(\uparrow42.76)}$ & $72.36_{(\downarrow22.72)}$ & $94.21_{(\downarrow0.48)}$ & $99.66_{(\uparrow15.09)}$ & $77.29_{(\uparrow19.89)}$ & $85.55_{(\uparrow3.18)}$ & $91.10_{(\downarrow8.9)}$ & $92.14_{(\downarrow7.58)}$ & $52.31_{(\downarrow45.96)}$ & $67.82_{(\downarrow32.05)}$ & $87.35_{(\downarrow1.75)}$ & $99.84_{(\uparrow9.67)}$ & $84.22_{(\uparrow29.67)}$ & $86.36_{(\downarrow1.53)}$ \\
Entropy         & $68.52_{(\downarrow29.82)}$ & $98.29_{(\uparrow2.87)}$ & $33.30_{(\downarrow5.17)}$ & $91.85_{(\downarrow3.23)}$ & $84.73_{(\downarrow9.96)}$ & $96.95_{(\uparrow12.38)}$ & $38.23_{(\downarrow19.17)}$ & $83.53_{(\uparrow1.16)}$ & $94.40_{(\downarrow5.6)}$ & $99.87_{(\uparrow0.15)}$ & $87.14_{(\downarrow11.13)}$ & $98.22_{(\downarrow1.65)}$ & $68.21_{(\downarrow20.89)}$ & $99.39_{(\uparrow9.22)}$ & $58.92_{(\uparrow4.37)}$ & $94.70_{(\uparrow6.81)}$ \\
Rank            & $71.97_{(\downarrow26.37)}$ & $87.54_{(\downarrow7.88)}$ & $29.97_{(\downarrow8.5)}$ & $85.73_{(\downarrow9.35)}$ & $98.66_{(\uparrow3.97)}$ & $96.36_{(\uparrow11.79)}$ & $58.52_{(\uparrow1.12)}$ & $92.58_{(\uparrow10.21)}$ & $86.86_{(\downarrow13.14)}$ & $94.56_{(\downarrow5.16)}$ & $68.35_{(\downarrow29.92)}$ & $93.33_{(\downarrow6.54)}$ & $92.34_{(\uparrow3.24)}$ & $92.49_{(\uparrow2.32)}$ & $65.19_{(\uparrow10.64)}$ & $87.16_{(\downarrow0.73)}$ \\
Log-Rank       & $83.37_{(\downarrow14.97)}$ & $98.75_{(\uparrow3.33)}$ & $32.40_{(\downarrow6.07)}$ & $91.19_{(\downarrow3.89)}$ & $95.07_{(\uparrow0.38)}$ & $98.89_{(\uparrow14.32)}$ & $47.05_{(\downarrow10.35)}$ & $88.34_{(\uparrow5.97)}$ & $99.57_{(\downarrow0.43)}$ & $99.96_{(\uparrow0.24)}$ & $94.09_{(\downarrow4.18)}$ & $99.03_{(\downarrow0.84)}$ & $87.92_{(\downarrow1.18)}$ & $99.31_{(\uparrow9.14)}$ & $67.70_{(\uparrow13.15)}$ & $95.02_{(\uparrow7.13)}$ \\
Log-Likelihood & $81.29_{(\downarrow17.05)}$ & $98.30_{(\uparrow2.88)}$ & $32.32_{(\downarrow6.15)}$ & $90.15_{(\downarrow4.93)}$ & $94.08_{(\downarrow0.61)}$ & $98.20_{(\uparrow13.63)}$ & $47.32_{(\downarrow10.08)}$ & $86.77_{(\uparrow4.4)}$ & $99.37_{(\downarrow0.63)}$ & $99.94_{(\uparrow0.22)}$ & $93.79_{(\downarrow4.48)}$ & $98.70_{(\downarrow1.17)}$ & $88.18_{(\downarrow0.92)}$ & $99.14_{(\uparrow8.97)}$ & $70.36_{(\uparrow15.81)}$ & $94.73_{(\uparrow6.84)}$ \\
DetectGPT       & $53.71_{(\downarrow44.63)}$ & $42.39_{(\downarrow53.03)}$ & $31.05_{(\downarrow7.42)}$ & $57.51_{(\downarrow37.57)}$ & $63.15_{(\downarrow31.54)}$ & $54.10_{(\downarrow30.47)}$ & $32.55_{(\downarrow24.85)}$ & $86.05_{(\uparrow3.68)}$ & $78.68_{(\downarrow21.32)}$ & $73.45_{(\downarrow26.27)}$ & $47.36_{(\downarrow50.91)}$ & $89.30_{(\downarrow10.57)}$ & $87.74_{(\downarrow1.36)}$ & $80.95_{(\downarrow9.22)}$ & $43.36_{(\downarrow11.19)}$ & $97.07_{(\uparrow9.18)}$ \\
Fast-DetectGPT  & $96.23_{(\downarrow2.11)}$ & $87.79_{(\downarrow7.63)}$ & $46.32_{(\uparrow7.85)}$ & $69.19_{(\downarrow25.9)}$ & $90.38_{(\downarrow4.31)}$ & $77.55_{(\downarrow7.02)}$ & $71.91_{(\uparrow14.51)}$ & $58.81_{(\downarrow23.56)}$ & $99.89_{(\downarrow0.11)}$ & $99.65_{(\downarrow0.07)}$ & $98.22_{(\downarrow0.05)}$ & $98.59_{(\downarrow1.28)}$ & $89.63_{(\uparrow0.53)}$ & $74.86_{(\downarrow15.31)}$ & $65.24_{(\uparrow10.9)}$ & $53.50_{(\downarrow34.3)}$ \\
Binoculars     & 98.34 & 95.42 & 38.47 & 95.08 & 94.69 & 84.57 & 57.40 & 82.37 & \textbf{100.00} & 99.72 & 98.27 & 99.87 & 89.10 & 90.17 & 54.55 & 87.89 \\
\midrule
DivScore (med) & \underline{99.86}$_{(\uparrow1.52)}$ & $99.07_{(\uparrow3.65)}$ & \textbf{99.17}$_{(\uparrow60.7)}$ & $99.27_{(\uparrow4.19)}$ & \underline{99.99}$_{(\uparrow5.3)}$ & \underline{99.87}$_{(\uparrow15.3)}$ & \textbf{99.45}$_{(\uparrow42.05)}$ & \underline{99.76}$_{(\uparrow17.39)}$ & 99.97$_{(\downarrow0.03)}$ & \underline{99.87}$_{(\uparrow0.15)}$ & $98.51_{(\uparrow0.24)}$ & $99.93_{(\uparrow0.06)}$ & \underline{99.98}$_{(\uparrow10.88)}$ & \underline{99.98}$_{(\uparrow9.81)}$ & $99.28_{(\uparrow44.73)}$ & \underline{99.99}$_{(\uparrow12.1)}$ \\
DivScore (law) & \underline{99.86}$_{(\uparrow1.52)}$ & \underline{99.61}$_{(\uparrow4.19)}$ & $98.65_{(\uparrow60.18)}$ & \underline{99.54}$_{(\uparrow4.46)}$ & \underline{99.99}$_{(\uparrow5.3)}$ & $99.81_{(\uparrow15.24)}$ & \underline{99.25}$_{(\uparrow41.85)}$ & \textbf{99.87}$_{(\uparrow17.5)}$ & $99.97_{(\downarrow0.03)}$ & $99.81_{(\uparrow0.09)}$ & \textbf{98.61}$_{(\uparrow0.34)}$ & \underline{99.94}$_{(\uparrow0.07)}$ & $99.94_{(\uparrow10.84)}$ & $99.93_{(\uparrow9.76)}$ & \underline{99.37}$_{(\uparrow44.82)}$ & \underline{99.99}$_{(\uparrow12.1)}$ \\
DivScore      & \textbf{99.87}$_{(\uparrow1.53)}$ & \textbf{99.90}$_{(\uparrow4.48)}$ & \underline{98.88}$_{(\uparrow60.41)}$ & \textbf{99.84}$_{(\uparrow4.76)}$ & \textbf{100.00}$_{(\uparrow5.31)}$ & \textbf{99.96}$_{(\uparrow15.39)}$ & $99.20_{(\uparrow41.8)}$  & \textbf{99.87}$_{(\uparrow17.5)}$ & \underline{99.99}$_{(\downarrow0.01)}$ & \textbf{99.99}$_{(\uparrow0.27)}$ & \underline{98.54}$_{(\uparrow0.27)}$ & \textbf{99.99}$_{(\uparrow0.12)}$ & \textbf{99.99}$_{(\uparrow10.89)}$ & \textbf{100.00}$_{(\uparrow9.83)}$ & \textbf{99.56}$_{(\uparrow45.01)}$ & \textbf{100.00}$_{(\uparrow12.11)}$ \\

\midrule
\multicolumn{1}{c}{} &\multicolumn{16}{c}{TPR @ 0.1\% FPR} \\
\midrule
RoB-base        & $0.00_{(\downarrow68.4)}$ & $35.40_{(\uparrow30.3)}$ & $0.10_{(\downarrow0.4)}$ & $0.00_{(\downarrow27.7)}$ & $0.00_{(\downarrow14.3)}$ & $62.60_{(\uparrow62.6)}$ & $0.10_{(\downarrow0.1)}$ & $0.10_{(\downarrow0.3)}$ & $0.00_{(\downarrow99.7)}$ & $36.90_{(\uparrow1.8)}$ & $0.00_{(\downarrow74.5)}$ & $0.20_{(\downarrow84.8)}$ & $0.04_{(\downarrow8.26)}$ & $59.20_{(\uparrow59.2)}$ & $0.20_{(\uparrow0.2)}$ & $0.10_{(\downarrow0.2)}$  \\
RoB-large      & $9.80_{(\downarrow58.6)}$ & $26.70_{(\uparrow21.6)}$ & 1.90$_{(\uparrow1.4)}$ & $0.70_{(\downarrow27.0)}$ & $5.08_{(\downarrow9.22)}$ & $78.80_{(\uparrow78.8)}$ & $0.90_{(\uparrow0.7)}$ & $0.30_{(\downarrow0.1)}$ & $0.94_{(\downarrow98.76)}$ & $23.20_{(\downarrow11.9)}$ & $0.10_{(\downarrow74.4)}$ & $0.30_{(\downarrow84.7)}$ & $0.62_{(\downarrow7.68)}$ & $82.30_{(\uparrow82.3)}$ & $2.80_{(\uparrow2.8)}$ & $1.30_{(\uparrow1.0)}$  \\
Entropy         & $0.00_{(\downarrow68.4)}$ & $28.80_{(\uparrow23.7)}$ & $0.10_{(\downarrow0.4)}$ & $0.00_{(\downarrow27.7)}$ & $0.60_{(\downarrow13.7)}$ & $10.20_{(\uparrow10.2)}$ & $0.00_{(\downarrow0.2)}$ & $0.00_{(\downarrow0.4)}$ & $5.90_{(\downarrow93.8)}$ & $89.20_{(\uparrow54.1)}$ & $4.40_{(\downarrow70.1)}$ & $0.00_{(\downarrow85.0)}$ & $0.00_{(\downarrow8.3)}$ & $37.10_{(\uparrow37.1)}$ & $0.10_{(\uparrow0.1)}$ & $0.00_{(\downarrow0.3)}$ \\
Rank            & $13.50_{(\downarrow54.9)}$ & $9.80_{(\uparrow4.7)}$ & $0.00_{(\downarrow0.5)}$ & $5.10_{(\downarrow22.6)}$ & $34.30_{(\uparrow20.0)}$ & $14.20_{(\uparrow14.2)}$ & $0.00_{(\downarrow0.2)}$ & $5.30_{(\uparrow4.9)}$ & $33.30_{(\downarrow66.4)}$ & $28.90_{(\downarrow6.2)}$ & $2.80_{(\downarrow71.7)}$ & $19.50_{(\downarrow65.5)}$ & $14.40_{(\uparrow6.1)}$ & $3.20_{(\uparrow3.2)}$ & $0.00_{(\uparrow0)}$ & $0.40_{(\uparrow0.1)}$ \\
Log-Rank        & $0.90_{(\downarrow67.5)}$ & $59.90_{(\uparrow54.8)}$ & $0.10_{(\downarrow0.4)}$ & $0.00_{(\downarrow27.7)}$ & $4.30_{(\downarrow10.0)}$ & $48.70_{(\uparrow48.7)}$ & $0.00_{(\downarrow0.2)}$ & $0.00_{(\downarrow0.4)}$ & $66.30_{(\downarrow33.4)}$ & $98.00_{(\uparrow62.9)}$ & $4.40_{(\downarrow70.1)}$ & $0.00_{(\downarrow85.0)}$ & $1.10_{(\downarrow7.2)}$ & $63.20_{(\uparrow63.2)}$ & $0.10_{(\uparrow0.1)}$ & $0.00_{(\downarrow0.3)}$ \\
Log-Likelihood  & $0.20_{(\downarrow68.2)}$ & $42.00_{(\uparrow36.9)}$ & $0.10_{(\downarrow0.4)}$ & $0.00_{(\downarrow27.7)}$ & $1.50_{(\downarrow12.8)}$ & $23.40_{(\uparrow23.4)}$ & $0.00_{(\downarrow0.2)}$ & $0.00_{(\downarrow0.4)}$ & $40.20_{(\downarrow59.5)}$ & $96.90_{(\uparrow61.8)}$ & $4.40_{(\downarrow70.1)}$ & $0.00_{(\downarrow85.0)}$ & $0.30_{(\downarrow8.0)}$ & $48.60_{(\uparrow48.6)}$ & $0.10_{(\uparrow0.1)}$ & $0.00_{(\downarrow0.3)}$ \\
DetectGPT       & $0.00_{(\downarrow68.4)}$ & $0.00_{(\downarrow5.1)}$ & $0.00_{(\downarrow0.5)}$ & $0.00_{(\downarrow27.7)}$ & $0.00_{(\downarrow14.3)}$ & $0.00_{(\uparrow0)}$ & $0.00_{(\downarrow0.2)}$ & $0.20_{(\downarrow0.2)}$ & $0.00_{(\downarrow99.7)}$ & $0.00_{(\downarrow35.1)}$ & $0.00_{(\downarrow74.5)}$ & $0.10_{(\downarrow84.9)}$ & $0.00_{(\downarrow8.3)}$ & $0.00_{(\uparrow0)}$ & $0.00_{(\uparrow0)}$ & $2.70_{(\uparrow2.4)}$  \\
Fast-DetectGPT  & $42.20_{(\downarrow26.2)}$ & $7.30_{(\uparrow2.2)}$ & $0.90_{(\uparrow0.4)}$ & $1.70_{(\downarrow26.0)}$ & $12.50_{(\downarrow1.8)}$ & $1.50_{(\uparrow1.5)}$ & $0.80_{(\uparrow0.6)}$ & $0.00_{(\downarrow0.4)}$ & $92.70_{(\downarrow7.0)}$ & $82.20_{(\uparrow47.1)}$ & $73.80_{(\downarrow0.7)}$ & $60.90_{(\downarrow24.1)}$ & $7.70_{(\downarrow0.6)}$ & $1.60_{(\uparrow1.6)}$ & $0.50_{(\uparrow0.5)}$ & $0.00_{(\downarrow0.3)}$ \\
Binoculars     & 68.40 & 5.10 & 0.50 & 27.70 & 14.30 & 0.00 & 0.20 & 0.40 & \textbf{99.70} & 35.10 & 74.50 & 85.00 & 8.30 & 0.00 & 0.00 & 0.30 \\
\midrule
DivScore (med)  & \textbf{94.10$_{(\uparrow25.7)}$} & \underline{71.80}$_{(\uparrow66.7)}$ & \textbf{81.00$_{(\uparrow80.5)}$} & $72.00_{(\uparrow44.3)}$ & \underline{99.20}$_{(\uparrow84.9)}$ & \underline{95.90}$_{(\uparrow95.9)}$ & \textbf{71.10$_{(\uparrow70.9)}$} & $83.70_{(\uparrow83.3)}$ & $98.30_{(\downarrow1.4)}$ & \underline{91.10}$_{(\uparrow56.0)}$ & $77.00_{(\uparrow2.5)}$ & $94.70_{(\uparrow9.7)}$ & \underline{97.60}$_{(\uparrow89.3)}$ & \underline{98.70}$_{(\uparrow98.7)}$ & $88.30_{(\uparrow88.3)}$ & $99.20_{(\uparrow98.9)}$ \\
DivScore (law)  & \underline{91.20}$_{(\uparrow22.8)}$ & $51.40_{(\uparrow46.3)}$ & $66.60_{(\uparrow66.1)}$ & \underline{80.20}$_{(\uparrow52.5)}$ & $97.80_{(\uparrow83.5)}$ & $83.90_{(\uparrow83.9)}$ & \underline{48.10}$_{(\uparrow47.9)}$ & \underline{93.50}$_{(\uparrow93.1)}$ & $97.70_{(\downarrow2.0)}$ & $76.20_{(\uparrow41.1)}$ & \textbf{78.00$_{(\uparrow3.5)}$} & \underline{97.30}$_{(\uparrow12.3)}$ & $94.20_{(\uparrow85.9)}$ & $92.60_{(\uparrow92.6)}$ & \underline{89.70}$_{(\uparrow89.7)}$ & \underline{99.70}$_{(\uparrow99.4)}$ \\
DivScore        & $90.40_{(\uparrow22.0)}$ & \textbf{94.50$_{(\uparrow89.4)}$} & \underline{69.60}$_{(\uparrow69.1)}$ & \textbf{88.00$_{(\uparrow60.3)}$} & \textbf{99.60}$_{(\uparrow85.3)}$ & \textbf{99.50$_{(\uparrow99.5)}$} & $42.80_{(\uparrow42.6)}$ & \textbf{96.50}$_{(\uparrow96.1)}$ & \underline{98.80}$_{(\downarrow0.9)}$ & \textbf{99.10}$_{(\uparrow64.0)}$ & \underline{77.30}$_{(\uparrow2.8)}$ & \textbf{98.10}$_{(\uparrow13.1)}$ & \textbf{98.80$_{(\uparrow90.5)}$} & \textbf{99.90}$_{(\uparrow99.9)}$ & \textbf{89.90}$_{(\uparrow89.9)}$ & \textbf{99.90$_{(\uparrow99.6)}$} \\

\bottomrule
\end{tabular}
}
\caption{Main Experiment - Detection AUROC score and TPR (at 0.1\% FPR) of baseline methods and DivScore on medical (MIMIC, PubMedQA) and legal datasets (OALC, LawStack). Methods are benchmarked against the Binoculars detector, with arrows indicating performance differences (↑/↓) relative to Binoculars. \textbf{Bold} indicates the best result, and \underline{underline} denotes the second-best.}
\vspace{0.8em}
\label{table: main}
\end{table*}

\begin{table*}[ht]
\small
\centering
\resizebox{\linewidth}{!}{%
\setlength{\tabcolsep}{4.5pt}
\renewcommand{\arraystretch}{1.15}
\begin{tabular}{l|ccc|ccc|ccc|ccc}
\toprule
\textit{Datasets} 
& \multicolumn{3}{c}{MIMIC} 
& \multicolumn{3}{c}{PubMedQA} 
& \multicolumn{3}{c}{OALC} 
& \multicolumn{3}{c}{LawStack} \\
\cmidrule(lr){2-4} \cmidrule(lr){5-7} \cmidrule(lr){8-10} \cmidrule(lr){11-13} 
\textit{Attacks} & w/o  & Paraphrase & Substitute 
& w/o  & Paraphrase & Substitute 
& w/o  & Paraphrase & Substitute 
& w/o  & Paraphrase & Substitute \\
\midrule
\multicolumn{1}{l}{} & \multicolumn{12}{c}{AUROC} \\
\midrule
RoB-base        & $66.79_{(\downarrow31.55)}$ & $47.08_{(\downarrow40.06)}$ & $55.01_{(\downarrow21.28)}$ & $85.79_{(\downarrow9.63)}$ & $68.59_{(\downarrow17.15)}$ & $66.86_{(\downarrow8.16)}$ & $58.33_{(\uparrow19.86)}$ & $41.92_{(\uparrow4.25)}$ & $40.87_{(\uparrow23.97)}$ & $65.10_{(\downarrow29.98)}$ & $51.62_{(\downarrow35.65)}$ & $49.83_{(\downarrow14.62)}$  \\
RoB-large      & $93.05_{(\downarrow5.29)}$ & $65.76_{(\downarrow21.38)}$ & $60.94_{(\downarrow15.35)}$ & $91.69_{(\downarrow3.73)}$ & $78.72_{(\downarrow7.02)}$ & $60.31_{(\downarrow14.71)}$ & \underline{81.23}$_{(\uparrow42.76)}$ & \underline{59.50}$_{(\uparrow21.83)}$ & \underline{45.04}$_{(\uparrow28.14)}$ & $72.36_{(\downarrow22.72)}$ & $48.53_{(\downarrow38.74)}$ & $35.68_{(\downarrow28.77)}$ \\
Entropy         & $68.52_{(\downarrow29.82)}$ & $9.84_{(\downarrow77.30)}$ & $23.88_{(\downarrow52.41)}$ & $98.29_{(\uparrow2.87)}$ & $82.56_{(\downarrow3.18)}$ & \underline{75.96}$_{(\uparrow0.94)}$ & $33.30_{(\downarrow5.17)}$ & $10.28_{(\downarrow27.39)}$ & $9.73_{(\downarrow7.17)}$ & $91.85_{(\downarrow3.23)}$ & $61.45_{(\downarrow25.82)}$ & $61.77_{(\downarrow2.68)}$\\
Rank           &  $71.97_{(\downarrow26.37)}$ & $57.54_{(\downarrow29.6)}$ & $11.82_{(\downarrow64.47)}$ & $87.54_{(\downarrow7.88)}$ & $82.75_{(\downarrow2.99)}$ & $52.59_{(\downarrow22.43)}$ & $29.97_{(\downarrow8.50)}$ & $27.23_{(\downarrow10.44)}$ & $9.65_{(\downarrow7.25)}$ & $85.73_{(\downarrow9.35)}$ & $72.22_{(\downarrow15.05)}$ & $43.18_{(\downarrow21.27)}$ \\
Log-Rank       & $83.37_{(\downarrow14.97)}$ & $18.42_{(\downarrow68.72)}$ & $25.53_{(\downarrow50.76)}$ & $98.75_{(\uparrow3.33)}$ & \underline{90.89}$_{(\uparrow5.15)}$ & $74.79_{(\downarrow0.23)}$ & $32.40_{(\downarrow6.07)}$ & $12.68_{(\downarrow24.99)}$ & $8.71_{(\downarrow8.19)}$ & $91.19_{(\downarrow3.89)}$ & $67.26_{(\downarrow20.01)}$ & $50.54_{(\downarrow13.91)}$ \\
Log-Likelihood  & $81.29_{(\downarrow17.05)}$ & $14.78_{(\downarrow72.36)}$ & $24.79_{(\downarrow51.50)}$ & $98.30_{(\uparrow2.88)}$ & $88.05_{(\uparrow2.31)}$ & $69.89_{(\downarrow5.13)}$ & $32.32_{(\downarrow6.15)}$ & $11.37_{(\downarrow26.30)}$ & $8.43_{(\downarrow8.47)}$ & $90.15_{(\downarrow4.93)}$ & $65.63_{(\downarrow21.64)}$ & $48.53_{(\downarrow15.92)}$ \\
DetectGPT       & $53.71_{(\downarrow44.63)}$ & $32.92_{(\downarrow54.22)}$ & $34.30_{(\downarrow41.99)}$ & $42.39_{(\downarrow53.03)}$ & $38.82_{(\downarrow46.92)}$ & $22.19_{(\downarrow52.83)}$ & $31.05_{(\downarrow7.42)}$ & $11.73_{(\downarrow25.94)}$ & $9.49_{(\downarrow7.41)}$ & $57.51_{(\downarrow37.57)}$ & $42.18_{(\downarrow45.09)}$ & $26.05_{(\downarrow38.40)}$\\
Fast-DetectGPT  & $96.24_{(\downarrow2.10)}$ & \textbf{87.82$_{(\uparrow0.68)}$} & \underline{78.01}$_{(\uparrow1.72)}$ & $87.77_{(\downarrow7.65)}$ & $79.99_{(\downarrow5.75)}$ & $39.75_{(\downarrow35.27)}$ & $46.32_{(\uparrow7.85)}$ & $48.69_{(\uparrow11.02)}$ & $19.30_{(\uparrow2.40)}$ & $69.19_{(\downarrow25.90)}$ & $70.46_{(\downarrow16.81)}$ & $26.39_{(\downarrow38.06)}$ \\
Binoculars     & \underline{98.34} & \underline{87.14} & 76.29        & 95.42 & 85.74 & 75.02 & 38.47        & 37.67        & 16.90        & \underline{95.08} & \underline{87.27} & \underline{64.45} \\
\midrule
DivScore       & \textbf{99.88$_{(\uparrow1.54)}$} & $78.74_{(\downarrow8.40)}$ & \textbf{97.03$_{(\uparrow20.74)}$} & \textbf{99.64$_{(\uparrow4.22)}$} & \textbf{93.85$_{(\uparrow8.11)}$} & \textbf{95.21$_{(\uparrow20.19)}$} & \textbf{98.88$_{(\uparrow60.41)}$} & \textbf{88.07$_{(\uparrow50.40)}$} & \textbf{94.63$_{(\uparrow77.73)}$} & \textbf{99.84$_{(\uparrow4.76)}$} & \textbf{91.29$_{(\uparrow4.02)}$} & \textbf{94.09$_{(\uparrow29.64)}$} \\
\midrule
\multicolumn{1}{l}{} & \multicolumn{12}{c}{TPR @ 0.1\% FPR} \\
\midrule
RoB-base       & $0.00_{(\downarrow68.40)}$ & $0.00_{(\downarrow11.30)}$ & $0.00_{(\downarrow8.10)}$ & \underline{35.40}$_{(\uparrow30.30)}$ & \underline{14.20}$_{(\uparrow13.80)}$ & \underline{14.50}$_{(\uparrow14.20)}$ & $0.10_{(\downarrow0.40)}$ & $0.00_{(\downarrow0.30)}$ & $0.00_{(\uparrow0.00)}$ & $0.00_{(\downarrow27.70)}$ & $0.00_{(\downarrow4.60)}$ & $0.00_{(\downarrow0.60)}$ \\
RoB-large       & $9.80_{(\downarrow58.60)}$ & $0.70_{(\downarrow10.60)}$ & $0.00_{(\downarrow8.10)}$ & $26.70_{(\uparrow21.60)}$ & $7.80_{(\uparrow7.40)}$ & $2.60_{(\uparrow2.30)}$ & \underline{1.90}$_{(\uparrow1.40)}$ & \underline{0.50}$_{(\uparrow0.20)}$ & \underline{0.40}$_{(\uparrow0.40)}$ & $0.70_{(\downarrow27.00)}$ & $0.20_{(\downarrow4.40)}$ & $0.00_{(\downarrow0.60)}$ \\
\midrule
Entropy         & $0.00_{(\downarrow68.40)}$ & $0.00_{(\downarrow11.30)}$ & $0.00_{(\downarrow8.10)}$ & $28.80_{(\uparrow23.70)}$ & $0.40_{(\uparrow0.00)}$ & $0.10_{(\downarrow0.20)}$ & $0.10_{(\downarrow0.40)}$ & $0.00_{(\downarrow0.30)}$ & $0.00_{(\uparrow0.00)}$ & $0.00_{(\downarrow27.70)}$ & $0.00_{(\downarrow4.60)}$ & $0.00_{(\downarrow0.60)}$ \\
Rank            & $13.50_{(\downarrow54.90)}$ & $0.90_{(\downarrow10.40)}$ & $0.00_{(\downarrow8.10)}$ & $9.80_{(\uparrow4.70)}$ & $1.50_{(\uparrow1.10)}$ & $0.00_{(\downarrow0.30)}$ & $0.00_{(\downarrow0.50)}$ & $0.00_{(\downarrow0.30)}$ & $0.00_{(\uparrow0.00)}$ & $5.10_{(\downarrow22.60)}$ & $0.40_{(\downarrow4.20)}$ & $0.00_{(\downarrow0.60)}$ \\
Log-Rank        & $0.90_{(\downarrow67.50)}$ & $0.00_{(\downarrow11.30)}$ & $0.00_{(\downarrow8.10)}$ & $59.90_{(\uparrow54.80)}$ & $5.60_{(\uparrow5.20)}$ & $0.80_{(\uparrow0.50)}$ & $0.10_{(\downarrow0.40)}$ & $0.00_{(\downarrow0.30)}$ & $0.00_{(\uparrow0.00)}$ & $0.00_{(\downarrow27.70)}$ & $0.00_{(\downarrow4.60)}$ & $0.00_{(\downarrow0.60)}$ \\
Log-Likelihood  & $0.20_{(\downarrow68.20)}$ & $0.00_{(\downarrow11.30)}$ & $0.00_{(\downarrow8.10)}$ & $42.00_{(\uparrow36.90)}$ & $1.10_{(\uparrow0.70)}$ & $0.00_{(\downarrow0.30)}$ & $0.10_{(\downarrow0.40)}$ & $0.00_{(\downarrow0.30)}$ & $0.00_{(\uparrow0.00)}$ & $0.00_{(\downarrow27.70)}$ & $0.00_{(\downarrow4.60)}$ & $0.00_{(\downarrow0.60)}$ \\
DetectGPT       & $0.00_{(\downarrow68.40)}$ & $0.00_{(\downarrow11.30)}$ & $0.00_{(\downarrow8.10)}$ & $0.00_{(\downarrow5.10)}$ & $0.00_{(\downarrow0.40)}$ & $0.00_{(\downarrow0.30)}$ & $0.00_{(\downarrow0.50)}$ & $0.00_{(\downarrow0.30)}$ & $0.00_{(\uparrow0.00)}$ & $0.00_{(\downarrow27.70)}$ & $0.00_{(\downarrow4.60)}$ & $0.00_{(\downarrow0.60)}$ \\
Fast-DetectGPT  & $41.80_{(\downarrow26.60)}$ & $5.90_{(\downarrow5.40)}$ & $4.60_{(\downarrow3.50)}$ & $6.90_{(\uparrow1.80)}$ & $2.10_{(\uparrow1.70)}$ & $0.00_{(\downarrow0.30)}$ & $0.90_{(\uparrow0.40)}$ & $0.00_{(\downarrow0.30)}$ & $0.00_{(\uparrow0.00)}$ & $1.70_{(\downarrow26.00)}$ & $1.50_{(\downarrow3.10)}$ & $0.00_{(\downarrow0.60)}$\\
Binoculars     & \underline{68.40} & \underline{11.30} & \underline{8.10} & 5.10 & 0.40 & 0.30 & 0.50 & 0.30 & 0.00 & \underline{27.70} & \underline{4.60} & \underline{0.60} \\
\midrule
DivScore  & \textbf{96.40$_{(\uparrow28.00)}$} & \textbf{12.80$_{(\uparrow1.50)}$} & \textbf{43.40$_{(\uparrow35.30)}$} & \textbf{85.80$_{(\uparrow80.70)}$} & \textbf{28.80$_{(\uparrow28.40)}$} & \textbf{23.00$_{(\uparrow22.70)}$} & \textbf{69.60$_{(\uparrow69.10)}$} & \textbf{3.40$_{(\uparrow3.10)}$} & \textbf{11.20$_{(\uparrow11.20)}$} & \textbf{88.00$_{(\uparrow60.30)}$} & \textbf{6.10$_{(\uparrow1.50)}$} & \textbf{12.30$_{(\uparrow11.70)}$} \\
\bottomrule
\end{tabular}%
}
\caption{Detector Robustness - Detection AUROC score and TPR (0.1\% FPR) for baselines and DivScore on GPT-4o-generated datasets with and without Adversarial Attacks (Paraphrase, Word Substitution). ``w/o'' indicates datasets without attacks. Methods are benchmarked against the Binoculars (SOTA), with arrows indicating performance differences ($\uparrow$/$\downarrow$) relative to Binoculars. \textbf{Bold} indicates the best result, and \underline{underline} denotes the second-best.}
\label{table:robust}
\vspace{0.8em}
\end{table*}
\section{Experiments}
In this section, we evaluate DivScore’s effectiveness in detecting machine-generated text within legal and medical domains. We further assess its robustness to adversarial attacks that attempt to bypass detection. Additionally, we analyse how domain knowledge distillation influences text scoring, and present ablation studies examining the effects of detector LLM selection, knowledge distillation, and scoring components on overall performance.

\subsection{Implementation Details }
This study investigates zero-shot, black-box detection settings that realistically reflect real-world deployments. In the zero-shot scenario, a detector must identify LLM-generated text without any task-specific labeled examples. The black-box constraint further withholds information about the source LLM and its logits, so detectors must rely on surrogate models for scoring. Together, these constraints mirror practical situations in which neither in-domain reference data nor the identity of the generating LLM is available.

\textbf{DivScore Detector}
DivScore employs Mistral-7B-v0.2~\cite{mistral7b} as the general LLM for text scoring and domain adaptation. The domain-adapted LLM is developed through a process of domain knowledge distillation, by finetuning on the Mistral-7B-v0.2. A combined knowledge distillation in both the medical and legal domains is conducted to optimize detector's capability within these specialized fields. In the main experiment, we additionally evaluate two variants of DivScore that undergo partial knowledge distillation in either the medical or legal domain alone, referred to as DivScore (med) and DivScore (law).

\textbf{Domain Knowledge Distillation }

We select DeepSeek-R1~\cite{deepseek_r1} as the teacher LLM due to its strong medical and legal expertise~\cite{deepseek_medical_reasoning,deepseek_legal_reasoning} and its permissive MIT license, which allows full knowledge distillation. Medical seed knowledge is sourced from 22,000 GPT-4o-generated, verifiable medical questions by~\citet{chen2024huatuogpto1medicalcomplexreasoning}. Legal seed knowledge is drawn from: (1) EQUALS~\cite{EQUALS}, with 6,909 legal questions from professional law consultations; (2) Open Legal QA~\cite{aus_legal_qa}, with 2,124 GPT-4-synthesized questions based on Australian legal cases; and (3) Pile-of-Law~\cite{pileoflaw}, with 15,393 legal questions from Reddit forums. Distillation knowledge is then generated by DeepSeek-R1 in the form of question answering. The resulting outputs are used to finetune the detector’s general LLM with Low Rank Adaptation (LoRA). Details for seed knowledge construction and LoRA finetuning are provided in Appendix~\ref{appendix:lora}. 

\textbf{Datasets }
We select four representative datasets in medical and legal domain to evaluate DivScore and baselines: (1) MIMIC-IV-Note (MIMIC)~\cite{mimic}, deidentified hospital notes for patients, (2) PubMedQA~\cite{pubmedqa},  questions and answers for biomedical research papers collected from PubMed, (3) Open Australia Legal Corpus (OALC)~\cite{aus_legal_qa}, a collection of Australian legislative and judicial documents, (4) LawStack~\cite{lawstack}, legal questions and answers collect from Stack Exchange. Within each dataset, we select 1,000 pairs of texts, each consisting of a human-written text and its LLM-generated counterparts. The source LLMs for LLM-generated text are chosen from two model families: GPT-4o~\cite{openai2024gpt4ocard}, O3-mini~\cite{o3mini}, DeepSeek-V3~\cite{deepseekai2025deepseekv3technicalreport} and DeepSeek-R1~\cite{deepseek_r1}, representing current SOTA LLMs available to the public. Detailed settings are provided in Appendix~\ref{section: dataset}.

\textbf{Baseline Detectors }
We benchmark DivScore against a comprehensive set of zero-shot and supervised detection methods. For zero-shot detectors, we include Log-Likelihood~\cite{loglik-roberta-solaiman2019release}, Entropy~\cite{entropy-lavergne2008detecting}, Rank, Log-Rank~\cite{rank_logrank-gehrmann2019gltr}, DetectGPT~\cite{detectgpt}, Fast-DetectGPT~\cite{fastdetectgpt}, and Binoculars~\cite{Binoculars}. For supervised detectors, we evaluate two RoBERTa-based classifiers (RoB-base and RoB-large) released by OpenAI~\cite{loglik-roberta-solaiman2019release}. In the zero-shot experiment setting, supervised detectors are assessed without further training on the current domain detection tasks. Detailed baseline definitions are provided in Appendix~\ref{section: baseline}.

\textbf{Evaluation Metrics }
Previous studies~\cite{detectllm,Text_Fluoroscopy,gec_score} mainly use AUROC to assess detector performance. However, AUROC, as an aggregate metric, may not reflect performance in low false positive regimes, which is crucial in specialized domains where misclassifying human-written texts as LLM-generated can be highly harmful. 
Therefore, in addition to AUROC, we report the True Positive Rate (TPR) at a False Positive Rate (FPR) of 0.1\%, following the evaluation protocol of \citet{Binoculars}. This metric reflects the TPR when, out of 1,000 human-written texts, at most one false positive is permitted.

\subsection{Main Results}
Table~\ref{table: main} shows the detection performance of baselines and DivScore on four medical and legal datasets generated by GPT-4o, O3-mini, DeepSeek-V3 and DeepSeek-R1, respectively. 
As shown, DivScore and its variants with partial knowledge distillation—DivScore (med) and DivScore (law)—demonstrate substantial performance gains over the Binoculars benchmark.
For DivScore (med), it yields on average a $14.3\%$ gain in AUROC and a $62.1$\% gain in TPR. For DivScore (law), it achieves a similar performance with $14.3$\% averaged gain in AUROC and $57.4$\% in TPR. The full DivScore, enhanced by distillation from both medical and legal domains, further pushes the average improvement to $14.4$\% in AUROC and $64.0$\% in TPR. These results underscore DivScore’s ability to detect specialized LLM-generated content with few false positives. Additionally, improvements of DivScore over its domain-specific variants indicate its generalizability across different specialized domains, showing its potential in cross-domain specialized LLM-generated text detection. Appendix~\ref{section: appendix_mainexp} and Figure~\ref{fig: auroc} provide a more detailed analysis of detection performance and the true vs. false positive trade-off for all detectors.

\subsection{Detector Robustness}
Recent studies show that paraphrasing~\cite{sadasivan2023can, krishna2023paraphrasing} and word-level perturbations~\cite{peng2023hidding, liang2023mutation} can effectively evade current AI-text detectors by altering lexical and syntactic features. To assess DivScore’s robustness in real-world application, we employ paraphrase~\cite{krishna2023paraphrasing} and word-level substitution~\citet{peng2023hidding} attacks on the GPT-4o texts. As shown in Table~\ref{table:robust}, both attacks substantially reduce the performance of zero-shot and supervised baselines. In contrast, DivScore remains robust, achieving on average 22.8\% higher in AUROC and 29.5\% higher in TPR than Binoculars. This robustness is likely due to the proposed knowledge distillation, which aligns the detector with LLM knowledge in specialized domains. Since adversarial attacks must preserve semantic and domain-specific content for usability, DivScore can still effectively recognize LLM-generated text, maintaining high detection accuracy even under strong perturbations. Detailed settings are provided in Appendix~\ref{appendix: robustness}.

\begin{table}[t]
\centering
\resizebox{\linewidth}{!}{%
\begin{tabular}{@{}lcccccc@{}}
\toprule
 & \multicolumn{2}{c}{Mean Token Cross-Entropy} & \multicolumn{2}{c}{Mean Token Entropy} & \multicolumn{2}{c}{Detector Score} \\
\cmidrule(lr){2-3} \cmidrule(lr){4-5} \cmidrule(lr){6-7}
 $Mean \pm Std $ & LLM & Human & LLM & Human & LLM & Human \\
\midrule
\multicolumn{2}{l}{\textbf{w/o Distillation}} & & & & & \\
\quad MIMIC & 1.81 $\pm$ 0.17 & 1.97 $\pm$ 0.21 & 0.98 $\pm$ 0.11 & 1.24 $\pm$ 0.15 & 0.54 $\pm$ 0.04 & 0.63 $\pm$ 0.04 \\
\quad PubMed & 1.70 $\pm$ 0.25 & 2.74 $\pm$ 0.47 & 0.99 $\pm$ 0.17 & 2.07 $\pm$ 0.40 & 0.58 $\pm$ 0.07 & 0.75 $\pm$ 0.04 \\
\quad OALC & 1.81 $\pm$ 0.30 & 1.80 $\pm$ 0.38 & 1.12 $\pm$ 0.22 & 1.35 $\pm$ 0.29 & 0.61 $\pm$ 0.06 & 0.75 $\pm$ 0.05 \\
\quad LawStack & 1.87 $\pm$ 0.21 & 2.42 $\pm$ 0.37 & 1.13 $\pm$ 0.14 & 1.72 $\pm$ 0.31 & 0.61 $\pm$ 0.05 & 0.71 $\pm$ 0.07 \\
\quad Avg. & 1.79 $\pm$ 0.23 & 2.23 $\pm$ 0.36 & 1.05 $\pm$ 0.16 & 1.60 $\pm$ 0.29 & 0.59 $\pm$ 0.06 & 0.71 $\pm$ 0.05 \\
\midrule
\multicolumn{2}{l}{\textbf{w/ Distillation}} & & & & & \\
\quad MIMIC & 3.56 $\pm$ 0.29 & 2.80 $\pm$ 0.30 & 0.46 $\pm$ 0.07 & 0.88 $\pm$ 0.13 & 0.13 $\pm$ 0.03 & 0.31 $\pm$ 0.03 \\
\quad PubMed & 4.56 $\pm$ 0.16 & 4.42 $\pm$ 0.14 & 0.37 $\pm$ 0.08 & 1.21 $\pm$ 0.32 & 0.08 $\pm$ 0.02 & 0.28 $\pm$ 0.04 \\
\quad OALC & 4.03 $\pm$ 0.37 & 3.04 $\pm$ 0.31 & 0.44 $\pm$ 0.09 & 0.68 $\pm$ 0.18 & 0.11 $\pm$ 0.02 & 0.23 $\pm$ 0.04 \\
\quad LawStack & 5.19 $\pm$ 0.27 & 4.32 $\pm$ 0.32 & 0.36 $\pm$ 0.06 & 0.90 $\pm$ 0.20 & 0.07 $\pm$ 0.01 & 0.21 $\pm$ 0.05 \\
\quad Avg. & 4.33 $\pm$ 0.27 & 3.64 $\pm$ 0.27 & 0.41 $\pm$ 0.07 & 0.92 $\pm$ 0.21 & 0.10 $\pm$ 0.02 & 0.26 $\pm$ 0.03 \\
\bottomrule
\end{tabular}%
}
\caption{Statistics (mean ± standard deviation) of DivScore and its components: cross-entropy, entropy for each dataset. Statistics are computed across four text source LLMs, with and without domain knowledge distillation. The "Avg." row shows dataset-averaged values.}
\label{table:distill_analysis}
\end{table}

\begin{figure}[t]
    \centering
    \includegraphics[width=1\linewidth]{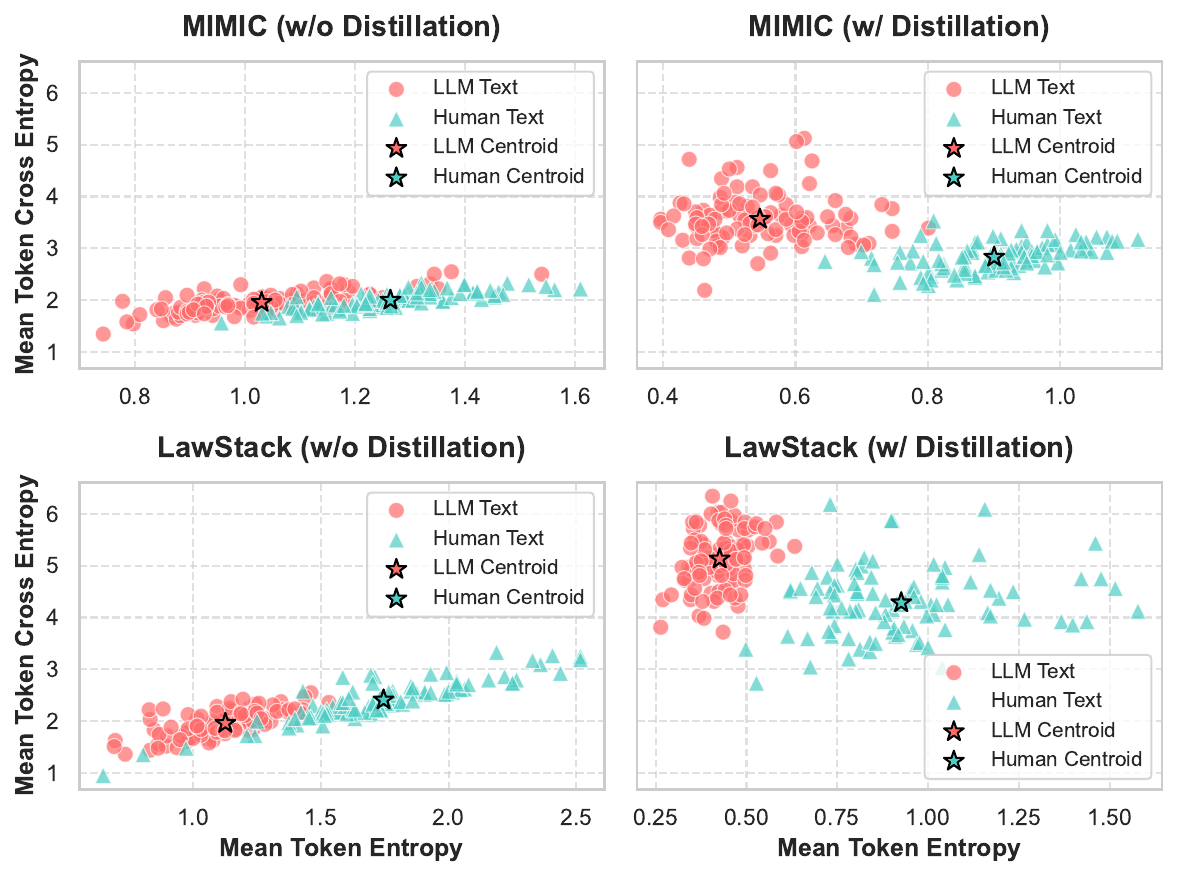}
    \caption{Text statistical signatures in DivScore for MIMIC and LawStack datasets with and without domain knowledge distillation. The data consists randomly selected 100 human-written and 100 GPT-4o-generated texts. Knowledge distillation greatly enlarges the cluster gap thereby enhances DivScore's performance. }
    \label{fig: w_wo_distill}
\end{figure}
\subsection{Knowledge Distillation Analysis}
\label{section: analysis_distill}
In this section, we examine how domain knowledge distillation influences the statistical signatures used by DivScore. We analyse the statistics of DivScore and its components - mean token entropy and mean token cross entropy with and without distillation. Mistral-7B-v0.2-Instruct, finetuned only on general instructions, serves as $\cal{M^*}$ for the "without distillation" baseline. As shown in Figure~\ref{fig: w_wo_distill}, knowledge distillation enhances the separability between LLM-generated and human-written texts, which is essential for accurate detection. Table~\ref{table:distill_analysis} additionally reports the mean and standard deviation of these signatures for both text types, giving the following findings.

\noindent \textbf{(1) Cross-Entropy \& Entropy.}
Knowledge distillation reduces the standard deviation of entropy (by 56.3\% for LLM texts and 27.6\% for human texts), making both clusters more compact. It also increases the centroid (mean) distance for cross-entropy by 36.2\%, enlarging the gap between LLM and human text clusters. These results, as shown in Table~\ref{table:distill_analysis}, confirm that knowledge distillation improves the separability of LLM- and human-generated texts. 
Cross-entropy statistics further support the hypothesis presented in Section~\ref{section: divscore}, namely that LLM-generated texts yield higher cross-entropy due to the disagreement between the domain-adapted LLM and the general-domain LLM. 

\noindent \textbf{(2) Detector Score.}
The proposed detector score, being the ratio of the above statistical signatures, shares a collective impact from knowledge distillation. As shown in Table~\ref{table:distill_analysis}, distillation leads to an overall 33.3\% increase in inter-group centroid (mean) distance, 66.7\% decrease in LLM score std and 40\% decrease in human score std. These results demonstrate that domain knowledge distillation enhances the distinguishability of LLM-generated texts by increasing the gap between LLM and human texts (centroid distance) and reducing intra-group variation (standard deviation). 

\subsection{Ablation Studies}
\label{sec:ablation}
\begin{figure}[t]
    \centering
    \includegraphics[width=\linewidth]{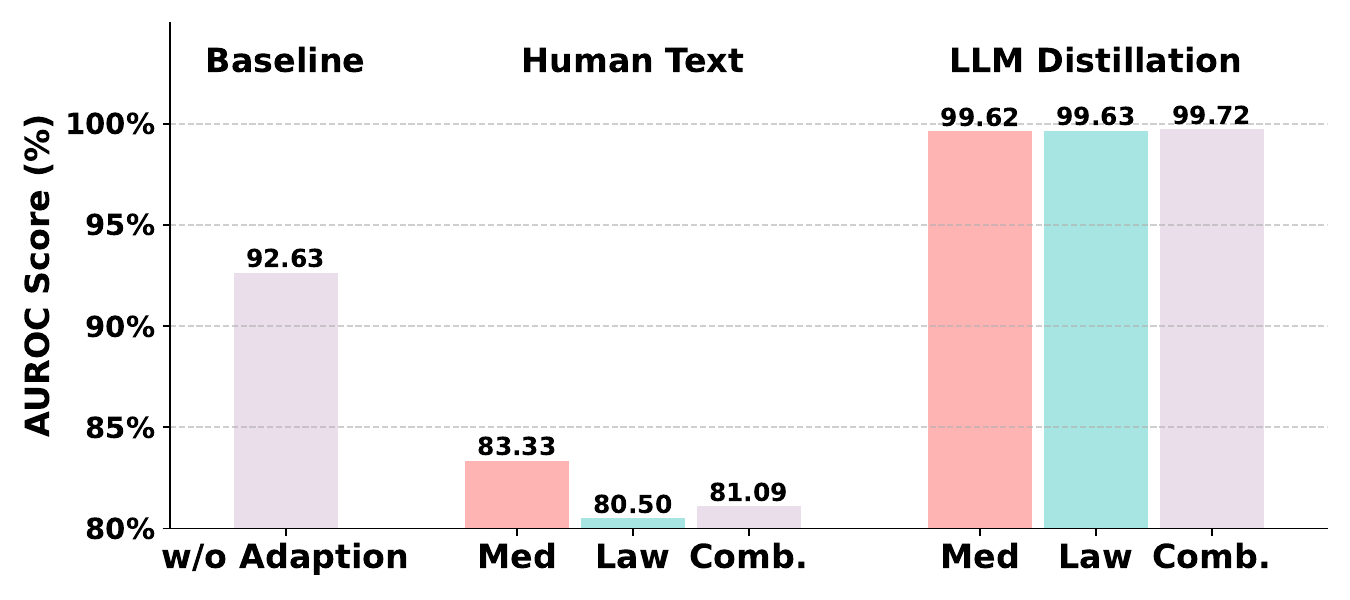}
    \caption{Comparison of detector AUROC scores across three settings: baseline (w/o domain adaptation), human domain text finetuning, and LLM domain knowledge distillation. For the later two settings, three variants are included: adaptation to medical, legal domain and their combination. AUROC scores are averaged across all evaluation datasets.}

    \label{fig: aucor_ablatinon}
\end{figure}
\begin{figure}[ht]
    \centering
    \includegraphics[width=\linewidth]{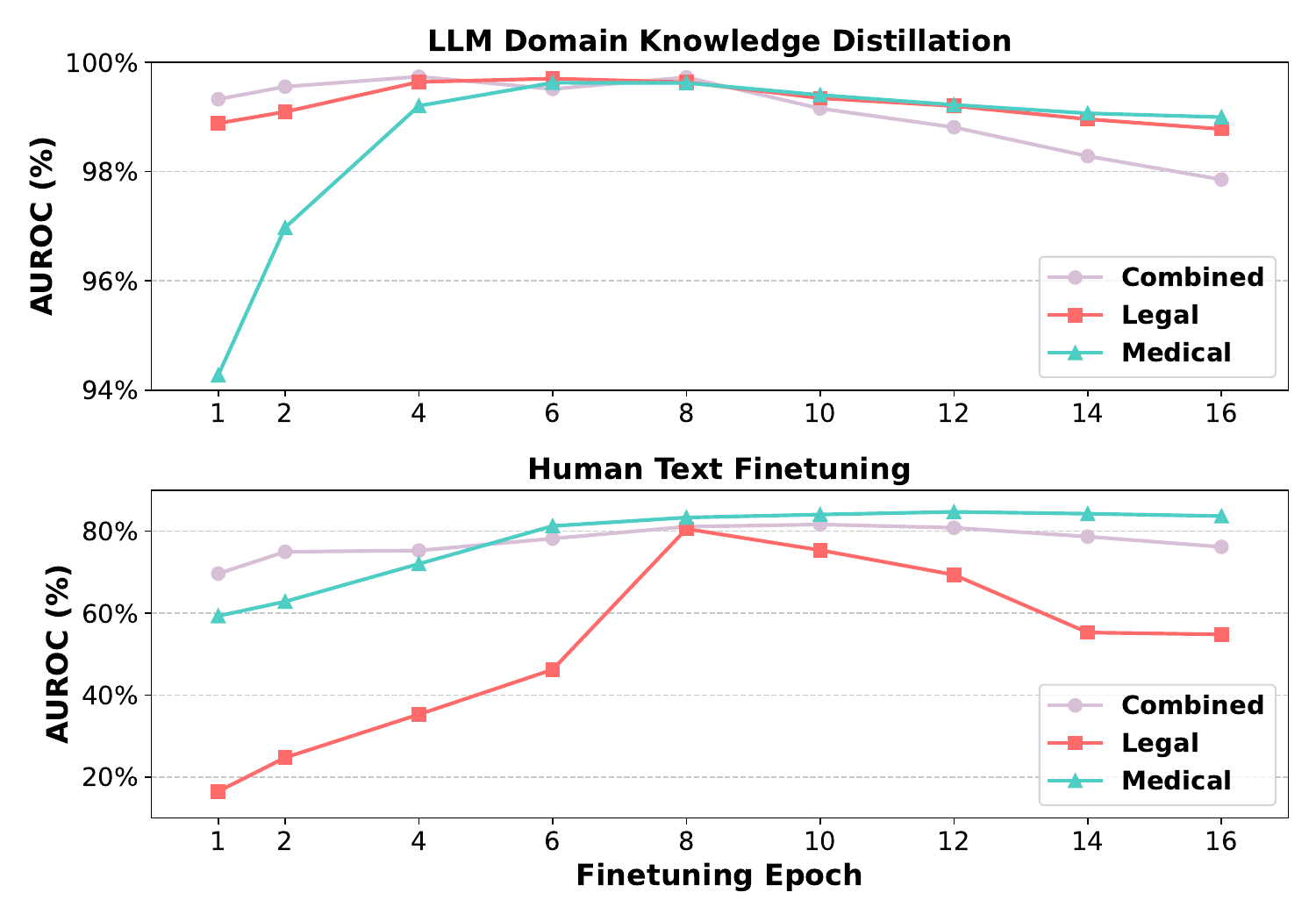}
    \caption{Detector performance (AUROC \%) of LLM distillation (proposed) and human text finetuning across finetuning epochs. AUROC scores are averaged across all evaluation datasets.}
    \label{fig: epoch_ablatinon}
\end{figure}

\textbf{Ablation on Domain Knowledge Distillation.}
In Section~\ref{section: method_distill}, we analysed the motivation for incorporating LLM domain knowledge distillation to improve DivScore's performance. We further empirically assess its impact on detection. Specifically, we compare DivScore with: (1) three variants where the domain adaptation for general LLM $\cal{M}$ is constructed by finetuning on human-written texts in medical, legal domain, and their combination, and (2) the baseline DivScore without domain adaptation, as referred in Section \ref{section: analysis_distill}.

    Figure~\ref{fig: aucor_ablatinon} presents the AUROC averaged across all datasets and text source LLMs. Incorporating LLM domain knowledge distillation significantly enhances detection performance, with respective AUROC improvements of 18.6\% and 7.1\% compared to DivScore variants adapted from human text corpus and the baseline. These findings suggest that LLM distillation is highly effective for constructing domain-adapted LLMs for DivScore, outperforming baseline and human text finetuning. We further investigated the impact of the number of finetuning epochs on the performance of each variant. As illustrated in Figure~\ref{fig: epoch_ablatinon}, all variants reach their peak or near-peak AUROC at epoch 8, after which performance declines. This observation supports our choice of finetuning epochs and ensures fair comparisons in the ablation studies. Intuitively, insufficient training leads to inadequate domain adaptation, while excessive finetuning may lead to forgetting of LLM linguistic characteristics in general domain, reducing detection effectiveness.

\begin{table}[t]
\centering
\resizebox{\linewidth}{!}{%
\large
\setlength{\tabcolsep}{2mm}{} 
\renewcommand{\arraystretch}{1.15}
\begin{tabular}{l|ccccc}
\toprule
\multirow{2}{*}{Methods} & \multicolumn{5}{c}{Text Source LLMs} \\
\cmidrule{2-6}
 & MIMIC & PubMedQA & OALC & LawStack & Avg \\
\midrule
Entropy (Mistral)               & 99.32 & 99.78 & 85.43 & 99.27 & 95.95 \\
Cross-Entropy (Mistral)         & 12.69 & 83.17 & 21.91 & 52.45 & 42.56 \\
DivScore (Mistral)       & \textbf{99.96} & \textbf{99.96} & \textbf{99.04} & \textbf{99.93} & \textbf{99.72} \\
\midrule
Entropy (Falcon)                  & 94.69 & 99.49 & 89.34 & 98.12 & 95.41\\
Cross-Entropy (Falcon)            & 25.37 & 73.19 & 54.26 & 46.58 & 49.85  \\
DivScore (Falcon)          & \textbf{99.70} & \textbf{99.13} & \textbf{95.87} & \textbf{99.13} & \textbf{98.46} \\
\midrule
Entropy (Qwen)                  & 98.81 & 99.88 & 89.03 & 99.38 & 96.78 \\
Cross-Entropy (Qwen)            & 20.86 & 94.97 & 19.73 & 61.36 & 49.23 \\
DivScore (Qwen)          & \textbf{99.96} & \textbf{94.43} & \textbf{97.45} & \textbf{99.33} & \textbf{97.79} \\
\midrule
Entropy (Llama)                 & 98.01 & 99.89 & 86.75 & 99.12 & 95.94 \\
Cross-Entropy (Llama)            & 10.31 & 64.55 & 17.10 & 32.23 & 31.05 \\
DivScore (Llama)         & \textbf{99.91} & \textbf{99.87} & \textbf{99.03} & \textbf{99.86} & \textbf{99.67} \\
\bottomrule
\end{tabular}%
}
\caption{Ablation results (AUROC \%) for DivScore and its scoring variants across four detector LLM architectures, showing that the proposed design consistently outperforms entropy or cross-entropy alone. AUROC scores are averaged across all text source LLMs.}
\label{table:base_llm}
\end{table}

\textbf{Ablation on Detector Components.}
We further validated DivScore’s design for two key detector components: the choice of LLM architecture and the effectiveness of the scoring mechanism. We compared four open-source LLMs: Falcon-7B~\cite{falcon}, Qwen2.5-8B~\cite{qwen2.5}, Llama3-8B~\cite{llama3}, and the proposed LLM, Mistral-7B-v0.2. We also evaluated each scoring component by comparing the overall detector score with its components: token entropy and token cross-entropy. The detector performance, averaged across all datasets and source LLMs (Table~\ref{table:base_llm}), yields three main findings: (1) DivScore achieves robust detection (Avg. AUROC > 95) across all LLM architectures, with Mistral-7B performing best; (2) the full detector scoring mechanism outperforms its components, confirming the necessity of the DivScore design. See detailed ablation settings in Appendix~\ref{section: ablation_1}.

\section{Conclusions}
In this work, we introduce DivScore, a theoretically grounded, zero-shot detector for identifying LLM-generated text in specialized and high-stakes domains such as medicine and law. Our analysis reveals that existing general zero-shot detectors suffer in these settings due to a distributional mismatch. To address this, DivScore leverages a normalized entropy-based scoring mechanism, augmented by domain knowledge distillation, to robustly detect LLM-generated specialized content. Empirical results on medical and legal domains show that DivScore consistently surpasses SOTA baselines, while remaining robust to adversarial attacks.

\section{Limitations}
We identify the following limitations of our work: (1) Domain Adaptation Requirement. Our approach requires tailored adaptation to target specialized domains. While the proposed detector demonstrates generalizability beyond a single domain, effective adaptation remains essential for high detection performance. This adaptation process may pose practical challenges, particularly in low-resource settings. Future work could focus on improving the efficiency and scalability of domain adaptation.
(2) Language Scope. Like most existing studies, our work focuses solely on detecting machine-generated text in English. The performance of DivScore in other languages, especially non-English and low-resource languages, remains unexamined. Expanding detection capabilities across languages—particularly in specialized and high-stakes contexts—is a critical direction for future research, including comprehensive benchmarking and adaptation strategies.

\bibliography{custom}

\appendix
\section{Proof of Theorem}
\label{section: theorem_proof}
We start the analysis with a standard entropy-based detector $d_{H}$, which utilizes an LLM (with generation distribution $Q$) to compute text entropy scores for classification. Given input medical text $X$, the detector's performance can be formalized as follows:

\begin{theorem_appendix}
Given a zero-shot LLM-generated text detector $D$ with detection score mapped from the detector LLM's text likelihood $p_{Q}(X)$ by any strictly monotonic function $f: \mathbb{R} \to \mathbb{R}$ :
\begin{equation}
D(X) = f(p_{Q}(X)),
\end{equation}
where $Q$ represents the text distribution calculated from $D$.
The performance of $D$, denoted as $\textup{AUROC}(D(X))$, satisfies:
\begin{equation}
\textup{AUROC}(D(X)) \propto^{+} \delta_{KL},
\end{equation}
where $\propto^+$ denotes positive monotonic association, and $\delta_{KL}$, effective KL divergence, is given as: 
\begin{equation}
\delta_{KL} =  D_{\text{KL}}(P \parallel Q') - D_{\text{KL}}(Q \parallel Q'),
\end{equation}
with $Q'$ indicating text distribution from the real source. $P$ denotes text distribution from human. $D_{KL}(\cdot||\cdot)$ denotes Kullback–Leibler(KL) divergence.
\end{theorem_appendix}
\begin{proof}
We begin the proof by considering the LLM entropy detector \( d_{H}(X) \) as a special case of $D(X)$. Detector \( d_{H} \) classifies sampled text \( x \) as machine-generated if its entropy under \( Q \) falls below a threshold \( \tau \):
\begin{equation}
d_{H}(x) = \begin{cases} 
1 & \text{if } H_{Q}(x) < \tau \\
0 & \text{otherwise}
\end{cases}
\end{equation}
where \( H_{Q}(x) \) denotes the LLM entropy of text $x$ sampled from $X$, based on LLM distribution \( Q \), given by:
\begin{equation}
    H_{Q}(x) = \sum_{x_i}p_Q(x_i | x_{<i}) \log p_Q(x_i | x_{<i})
\end{equation}

\medskip
\noindent\textbf{Distributional Assumptions:}

Assume the text variable $X$'s LLM entropy values follow normal distributions:
\begin{equation}
\label{eq:guassian}
\begin{aligned}
H_{Q}(X) & \sim \mathcal{N}(\mu_{Q'}, \sigma_{Q'}^2) \quad  \text{if } X \sim Q' \\
H_{Q}(X) &\sim \mathcal{N}(\mu_{P}, \sigma_{P}^2) \quad  \text{if } X \sim P
\end{aligned}
\end{equation}
with means derived from cross-entropy among detector distribution \( Q \), human distribution \( P \) and LLM text generation distribution \( Q' \):
\begin{equation}
\label{eqaution: mean}
\begin{aligned}
\mu_{Q} &= H(Q) + D_{\text{KL}}(Q \parallel Q') \\
\mu_{P} &= H(P) + D_{\text{KL}}(P \parallel Q')
\end{aligned}
\end{equation}

\medskip
\noindent\textbf{ROC Curve Construction:}

The True Positive Rate (TPR) and False Positive Rate (FPR) are defined as:
\begin{equation}
\begin{aligned}
\text{TPR} &= \mathbb{P}_{X \sim Q'}(H_{Q}(X) < \tau) \\
\text{FPR} &= \mathbb{P}_{X \sim P}(H_{Q}(X) < \tau)
\end{aligned}
\end{equation}
given Equation~\ref{eq:guassian},
\begin{equation}
\begin{aligned}
\text{TPR}(\tau) &= \Phi\left( \frac{\tau - \mu_{Q}}{\sigma_{Q'}} \right) \\
\text{FPR}(\tau) &= \Phi\left( \frac{\tau - \mu_{P}}{\sigma_{P}} \right)
\end{aligned}
\end{equation}
where \( \Phi(\cdot) \) is the CDF of the standard normal distribution.

\medskip
\noindent\textbf{AUROC Derivation:}

The AUROC is the integral of TPR over FPR, denote FPR as \( p \) for simplicity:
\begin{equation}
\label{equation: integral}
\text{AUROC} = \int_{0}^{1} \Phi\left( \Phi^{-1}(p) \frac{\sigma_{P}}{\sigma_{Q'}} + \frac{\delta_{KL}}{\sigma_{Q'}} \right) dp
\end{equation}
where \( \delta_{KL} = \mu_{P} - \mu_{Q} = D_{\text{KL}}(P\parallel Q') - D_{\text{KL}}(Q \parallel Q') \), given Equation~\ref{eqaution: mean}.

Let \( z = \Phi^{-1}(p) \), then \( dp = \phi(z)dz \), where \( \phi(z) \) is the Cumulative Density Function (CDF) of standard normal distribution. Equation~\ref{equation: integral} simplifies to:
\begin{equation}
\begin{aligned}
\text{AUROC} &= \int_{-\infty}^{\infty} \Phi\left( \frac{\sigma_{P}}{\sigma_{Q'}}z + \frac{\delta_{KL}}{\sigma_{Q'}} \right) \phi(z) dz \\
&= \Phi\left( \frac{\delta_{KL}}{\sqrt{\sigma_{P}^2 + \sigma_{Q'}^2}} \right)
\end{aligned}
\end{equation}

Since the CDF of normal distribution is monotonically increasing, then, for detector $d_{H}$, we have:
\begin{equation}
\text{AUROC}(d_{H}(X)) \propto^+ \delta_{KL}
\end{equation}
We further extend such conclusion to the complete definition of $D(X)$. 
Let $s(x) = f(p_Q(x))$, where $f: \mathbb{R} \to \mathbb{R}$ is any strictly monotonic function. For any $x_1, x_2 \in \mathcal{X}$:
\begin{equation}
\label{equation: monoton}
\resizebox{\linewidth}{!}{$
  H_Q(x_1) > H_Q(x_2) \iff 
  \begin{cases} 
    s(x_1) > s(x_2) & (f \nearrow) \\ 
    s(x_1) < s(x_2) & (f \searrow)
  \end{cases}
$}
\end{equation}

Equation~\ref{equation: monoton} implies $s(\cdot)$ preserves/reverses the exact ranking induced by $H_Q(\cdot)$.
For any  threshold $\tau_s$ for detector $s$, define equivalent $H_Q$ threshold:
\begin{equation}
\label{eq:threshold}
    \tau_H := H_Q(f^{-1}(\tau_s))
\end{equation}
Suppose $f \nearrow$, for any $\tau_s$:
\begin{equation}
\begin{aligned}
 \text{TPR}(\tau_s) &=  \mathbb{P}_{X \sim Q'}(s(X) < \tau_s)  \\
 \text{FPR}(\tau_s) &= \mathbb{P}_{X \sim P}(s(X) < \tau_s)
\end{aligned}
\end{equation}
According to Equation~\ref{eq:threshold}:
\begin{equation}
\begin{aligned}
 \text{TPR}(\tau_s) &=  \mathbb{P}_{X \sim Q'}(H_Q(X) < \tau_H) \\
 \text{FPR}(\tau_s) &= \mathbb{P}_{X \sim P}(H_Q(X) < \tau_H)
\end{aligned}
\end{equation}
Similarly, for $f \searrow$, for any $\tau_s$:
\begin{equation}
\begin{aligned}
 \text{TPR}(\tau_s) &= \mathbb{P}_{X \sim Q'}(H_Q(X) < \tau_H) \\
 \text{FPR}(\tau_s) &= \mathbb{P}_{X \sim P}(H_Q(X) < \tau_H)
\end{aligned}
\end{equation}

The ROC curves are therefore identical for detector \( s(X) \) and \( H_Q(X) \) in either settings, giving:
\begin{equation}
\text{AUROC}(D(X)) = \text{AUROC}(d_H(X))
\end{equation}
Therefore, we have
\begin{equation}
    \text{AUROC}(D(X)) \propto^+ \delta_{KL}.
\end{equation}
\end{proof}

\section{Experimental Settings}
This section describes the experimental configurations for our main evaluation, adversarial attacks, and ablation studies. Table~\ref{table:LLMs} provides detailed specifications of the large language models used throughout our tests. All experiments were performed on a Linux server with 8 NVIDIA H100-80G GPUs.

\begin{table*}[htbp]
\centering
\resizebox{\linewidth}{!}{%
\begin{tabular}{llll}
\toprule
\textbf{Model} & \textbf{Model File/Service} & \textbf{Parameters} & \textbf{Usage} \\ \midrule

GPT-4o~\cite{openai2024gpt4ocard} & OpenAI/gpt-4o & NA & Text Source \\ 
O3-mini~\cite{o3mini} & OpenAI/O3-mini & NA & Text Source\\
DeepSeek-R1~\cite{deepseek_r1} & deepseek-ai/DeepSeek-R1 & 671B (MoE) & Text Source \\ 
DeepSeek-V3~\cite{deepseekai2025deepseekv3technicalreport} & deepseek-ai/DeepSeek-V3 & 685B (MoE) & Text Source  \\ \midrule 
DIPPER~\cite{krishna2023paraphrasing} & kalpeshk2011/dipper-paraphraser-xxl & 11B & Paraphrase Attack \\
BERT-base~\cite{bert-base} & google-bert/bert-base-uncased & 110M & Word Substitution Attack \\ \midrule
Llama3~\cite{llama3} & meta-llama/Meta-Llama-3-8B & 8B &  Detector Model \\
Mistral-v0.2~\cite{mistral7b} & mistral-community/Mistral-7B-v0.2 & 7B &  Detector Model \\
Qwen2.5~\cite{qwen2.5} & Qwen/Qwen2.5-7B & 7B & Detector Model\\
Falcon~\cite{falcon} & tiiuae/falcon-7b & 7B & Detector Model\\
Falcon-Instruct~\cite{falcon} & tiiuae/falcon-7b-Instruct & 7B & Detector Model\\
T5~\cite{t5}& google-t5/t5-3b & 3B & Detector Model \\
GPT-Neo~\cite{GPT-Neo} & EleutherAI/gpt-neo-2.7B & 2.7B & Detector Model\\
GPT-J~\cite{gpt-j-6B}& EleutherAI/gpt-j-6B & 6B & Detector Model \\
RoBERTa-base~\cite{loglik-roberta-solaiman2019release} & openai-community/roberta-base-openai-detector & 110M & Detector Model \\
RoBERTa-large~\cite{loglik-roberta-solaiman2019release} & openai-community/roberta-large-openai-detector & 340M & Detector Model\\
\bottomrule
\end{tabular}%
}
\caption{Details of LLMs used in the experiments, including machine-generated text source models, adversarial attack models and detector models}
\label{table:LLMs}
\end{table*}

\begin{table}[ht]
    \centering
    \begin{tabular}{c}
        \begin{tcolorbox}[
            colback=gray!10,
            colframe=black,
            width=\linewidth,
            boxrule=0.8pt,
            arc=1mm,
            auto outer arc,
            left=2mm,
            right=2mm,
            top=2mm,
            bottom=2mm,
            breakable
        ]
        \setlength{\parindent}{0pt}
        \setlength{\parskip}{3pt}
[ \\
        \{"role": "system", "content": "Continue this hospital discharge summary."\}, \\
        \{"role": "user", "content": "<prefix: 1/3 document content> "\}\\
]
        \end{tcolorbox}
    \end{tabular}
    \caption{Prompt Template for MIMIC Dataset}
    \label{table: mimic_prompt}
\end{table}

\begin{table}[ht]
    \centering
    \begin{tabular}{c}
        \begin{tcolorbox}[
            colback=gray!10,
            colframe=black,
            width=\linewidth,
            boxrule=0.8pt,
            arc=1mm,
            auto outer arc,
            left=2mm,
            right=2mm,
            top=2mm,
            bottom=2mm,
            breakable
        ]
        \setlength{\parindent}{0pt}
        \setlength{\parskip}{3pt}
[ \\
        \{"role": "system", "content": "Continue to write this legal text."\}, \\
        \{"role": "user", "content": "<prefix: 1/3 document content> "\}\\
]
        \end{tcolorbox}
    \end{tabular}
    \caption{Prompt Template for OALC Dataset}
    \label{table: olac_prompt}
\end{table}

\subsection{Text Detection Datasets}
\label{section: dataset}
In this study, we focus on detecting LLM-generated text within written English. All detection texts are originally composed in English. The datasets are organized in two formats: (1) Documents: The MIMIC~\cite{mimic} and OALC~\cite{open-australian-legal-corpus} datasets consist of independent, human-written medical and legal documents; (2) QA pairs: The PubMedQA~\cite{pubmedqa} and LawStack~\cite{lawstack} datasets contain paired, human-written questions and answers. To compile our test set, we generate 1,000 samples for each dataset and each source model (GPT-4o, O3-mini, DeepSeek-V3, DeepSeek-R1) by calling API service.


For the document datasets, we generate corresponding LLM-generated texts by using the first one-third of each document’s tokens as a prompt and instructing the text source LLMs to continue the writing. To ensure a fair comparison, the human-written texts are also truncated, retaining only the remaining two-thirds of the original document. For the QA pair datasets, the human text consists of the human-authored answers, while the LLM-generated text is produced by prompting the questions to the text source LLMs. Specifically, we initiate the text generation process by sending the following prompt to the LLM API service, as shown in Table~\ref{table: mimic_prompt}, Table~\ref{table: olac_prompt} and Table~\ref{table: qa_prompt}.

\begin{table}[ht]
    \centering
    \begin{tabular}{c}
        \begin{tcolorbox}[
            colback=gray!10,
            colframe=black,
            width=\linewidth,
            boxrule=0.8pt,
            arc=1mm,
            auto outer arc,
            left=2mm,
            right=2mm,
            top=2mm,
            bottom=2mm,
            breakable
        ]
        \setlength{\parindent}{0pt}
        \setlength{\parskip}{3pt}
[ \\
        \{"role": "system", "content": "Answer this question truthfully:"\},\\
        \{"role": "user", "content": "<prefix: question>"\}\\
]
        \end{tcolorbox}
    \end{tabular}
    \caption{Prompt Template for PubMedQA and LawStack Dataset}
    \label{table: qa_prompt}
\end{table}

\subsection{Baseline Detectors}
\label{section: baseline}
In this section, we present the benchmark detectors and their configurations used in the main experiments. For zero-shot detectors, classification scores are obtained by extracting the logits from the detector’s surrogate language model (and its variants). Classical zero-shot methods—including Log-Likelihood, Rank, Log-Rank, and Entropy—employ Mistral-7B-v0.2 to ensure consistency with the DivScore detector LLM setup. Recent zero-shot detectors, such as DetectGPT, Fast-DetectGPT, and Binoculars, are implemented using the original LLM configurations as specified in their respective publications. For supervised detectors, we utilize the pre-trained models released by the original authors and apply them directly for out-of-distribution (OOD) text detection in our main experiments. Detailed descriptions of the baseline methods are provided below:
\\ \\
\textbf{Log-Likelihood} This zero-shot method uses a language model to compute the log-probability of each token in a given text~\cite{loglik-roberta-solaiman2019release}. Specifically, the averaged log-probability across all tokens is used as the detection score. A higher score corresponds to an increased probability of the text being generated by LLM.
\\ \\
\textbf{Rank} This zero-shot method utilizes the absolute ranking of tokens as determined by a language model~\cite{rank_logrank-gehrmann2019gltr}. By computing the mean rank score, a higher average value indicates an increased probability that the text is machine-generated.
\\ \\
\textbf{Log-Rank} An enhanced version of Rank method. This zero-shot method applies logarithmic transition to each token's absolute rank~\cite{rank_logrank-gehrmann2019gltr}. A higher mean log-rank value indicates a higher likelihood that the text is machine-generated. 
\\ \\
\textbf{Entropy} This zero-shot approach leverages a language model to calculate the mean entropy score across all tokens in a given text~\cite{entropy-lavergne2008detecting}. Since human-written content tends to exhibit greater unpredictability from the perspective of an LLM, a lower mean entropy score suggests a higher likelihood of the text being machine-generated.
\\ \\
\textbf{DetectGPT} A zero-shot method using probabilistic curvature for detection~\cite{detectgpt}. The approach involves perturbing text samples and analysing the resulting changes in a pretrained language model's log-likelihood. The underlying theory suggests that LLM-generated texts tend to occupy local optima in the LLM's log probability space, causing perturbations to mostly decrease their probability scores. In contrast, human-written texts does not follow such property. In our experiments, DetectGPT utilizes its default optimal setting in the literature: GPT-Neo-2.7B~\cite{GPT-Neo} as the surrogate model and T5-3B~\cite{t5} as mask-filling model.
\\ \\
\begin{table}[ht]
    \centering
    \begin{tabular}{c}
        \begin{tcolorbox}[
            colback=gray!10,
            colframe=black,
            width=\linewidth,
            boxrule=0.8pt,
            arc=1mm,
            auto outer arc,
            left=2mm,
            right=2mm,
            top=2mm,
            bottom=2mm,
            breakable
        ]
        \setlength{\parindent}{0pt}
        \setlength{\parskip}{3pt}
[ \\
    \{"role": "system","content": "You are a medical professional, reply the medical question with professional knowledge."\},\\
    \{"role": "user", "content": "<prefix: question>. Answer:"\}\\
]
        \end{tcolorbox}
    \end{tabular}
    \caption{Prompt Template for Medical Seed Knowledge}
    \label{table: med_distill}
\end{table}

\begin{table}[ht]
    \centering
    \begin{tabular}{c}
        \begin{tcolorbox}[
            colback=gray!10,
            colframe=black,
            width=\linewidth,
            boxrule=0.8pt,
            arc=1mm,
            auto outer arc,
            left=2mm,
            right=2mm,
            top=2mm,
            bottom=2mm,
            breakable
        ]
        \setlength{\parindent}{0pt}
        \setlength{\parskip}{3pt}
[ \\
    \{"role": "system","content": "You are a legal professional, reply the legal consult with professional knowledge."\},\\
    \{"role": "user", "content": "<prefix: question>. Answer:"\}\\
]
        \end{tcolorbox}
    \end{tabular}
    \caption{Prompt Template for Legal Seed Knowledge}
    \label{table: legal_distill}
\end{table}
\textbf{Fast-DetectGPT}~\citet{fastdetectgpt} proposed a new zero-shot approach to address issue of high computational cost in DetectGPT. It employs an LLM sampling method to replace the perturbation step of DetectGPT. Leveraging conditional probability curvature as metric, Fast-DetectGPT identifies the word-choice pattern differences in between LLM and human text. We chose the optimal settings reported in the literature, using GPT-Neo-2.7B as the scoring model and GPT-J-6B~\cite{gpt-j-6B} as the reference model.
\\ \\ 
\textbf{Binoculars}~\citet{Binoculars} proposes a zero-shot detection method that leverages a pair of LLMs to compute the ratio of a text’s perplexity to its cross-perplexity. This perplexity ratio measures how one model reacts to another’s token-level predictions, exploiting systematic differences in LLM and human-generated text. A lower score indicates higher likelihood of machine-generated origin, as LLM-produced text tends to be less surprising ``relative to the baseline perplexity of an LLM acting on the same string''~\citep{Binoculars}. In the main experiments, we chose the optimal detector settings reported in the literature, using Flacon-7B~\cite{falcon}, Flacon-7B-Instruct~\cite{falcon} as observer and performer model.
\\ \\
\textbf{OpenAI RoBERTa Detectors} A set of popular and competitive supervised detectors with RoBERTa architecture.~\citet{loglik-roberta-solaiman2019release} introduced a labelled dataset comprising GPT-2-generated outputs and WebText samples for supervised training on LLM-generated text detection. Leveraging this dataset, they finetuned two classifiers based on RoBERTa-base and RoBERTa-large model, achieving over 95\% accuracy on GPT-2 text detection. In the main experiments, we adopt the original model weights released by OpenAI and evaluate their OOD detection performance on LLM-generated medical and legal texts without additional training.

\begin{table}[t]
    \centering
    \begin{tabular}{c}
        \begin{tcolorbox}[
            colback=gray!10,
            colframe=black,
            width=\linewidth,
            boxrule=0.8pt,
            arc=1mm,
            auto outer arc,
            left=2mm,
            right=2mm,
            top=2mm,
            bottom=2mm,
            breakable
        ]
        \setlength{\parindent}{0pt}
        \setlength{\parskip}{3pt}
[ \\
    {"role": "system",
     "content": "You are a legal document translator."},\\
    {"role": "user", "content": "Translate this text into English: <question prefix>. Translation output:"}\\
]
        \end{tcolorbox}
    \end{tabular}
    \caption{Prompt Template for EQUALS legal question translation}
    \label{table: equlas_prompt}
\end{table}

\subsection{LLM Knowledge Distillation}
\label{appendix:lora}
\textbf{Domain Seed Knowledge}
To facilitate detector domain adoption, seed knowledge serves as a catalyst, prompting the teacher LLM to generate more elaborate and detailed domain knowledge based on its initial information. The prompt of seed knowledge is formulated in QA format and sent to DeepSeek-R1 through its official API service.

In this work, we focus on knowledge distillation using written English text. All seed knowledge datasets, except for EQUALS, are originally composed in English, and all prompted responses from DeepSeek-R1 are generated in English. For the EQUALS~\cite{EQUALS} dataset, whose original legal questions are in Chinese, we utilize DeepSeek-V3 to translate the questions into English prior to prompting the teacher LLM. Table~\ref{table: equlas_prompt} shows the prompt for translation.

\noindent \textbf{Distillation Knowledge base}
We construct the distillation knowledge base using the response content generated by the teacher LLM (DeepSeek-R1). Since the reasoning content produced by R1 reflects the model's internal self-reasoning rather than professional expert output, it is not utilized for detector domain adaptation. A summary of the medical and legal distillation knowledge bases is provided in Table~\ref{table: knowledge_base}.

\begin{table}[htbp]
    \centering
    \resizebox{\linewidth}{!}{
    \begin{tabular}{l p{0.45\linewidth} p{0.25\linewidth}}
        \toprule
        Knowledge Base & Seed Knowledge Composition &  Tokens Size\\
        \midrule
        DivScore (med) & Verifiable Medical Questions~\cite{chen2024huatuogpto1medicalcomplexreasoning} & 9.2M\\
        DivScore (law) & EQUALS~\cite{EQUALS}, Open Legal QA~\cite{aus_legal_qa}, Pile-of-Law~\cite{pileoflaw} & 17.3M \\
        DivScore & Combined from above & 26.5M\\
        \bottomrule
    \end{tabular}
    }
    \caption{Domain Adaptation: Distillation Knowledge Base Summary.  (M: Million Tokens)}
    \label{table: knowledge_base}
\end{table}

\noindent\textbf{Student Learning}
We enforce the student The LoRA hyperparameters were configured as LoRA rank: 64 and alpha scaling factor: 128. Supervised finetuning selectively updates the self-attention projection matrices (query, key, value) and feed-forward network components (gate, up projections) within the Mistral architecture. The finetuning is conducted with a bfloat16 parameter precision with number of training epoch set as 8, learning rate set as $2e^{-4}$, batch size set as 12 and maximum sequence length set as 2048.

\subsection{Detector Robustness}
\label{appendix: robustness}
To assess DivScore's robustness in real-world scenarios, we apply two types of adversarial attacks to the dataset. For the paraphrase attack, we utilize the DIPPER paraphraser introduced by~\citet{krishna2023paraphrasing}, an 11B-parameter T5-based LLM finetuned for paragraph-level paraphrasing with controllable perturbation strength, determined by lexical diversity and content reordering. DIPPER receives the candidate text along with these two parameters as input and generates paraphrased outputs at the specified perturbation level. In our experiments, both the lexical diversity and content reordering factors are set to 20\%, representing a moderate attack intensity.

For the word substitution attack, we follow the protocol of~\citet{peng2023hidding}, selecting the most frequent words in the text for semantically coherent replacement using a BERT-base model~\cite{bert-base}. Specifically, the BERT model operates at its maximum sequence length (512 tokens), replacing the top 20\% most frequent words within a 50-token context window. Common stop-words, which lack substantial semantic content, are excluded from substitution.

\subsection{Ablation Study}
\label{section: ablation_1}
In the ablation study on domain knowledge distillation, we benchmark the proposed domain adaptation approach with human text finetuning. For human text finetuning, we select medical and legal corpus to conduct the same LoRA finetuning as knowledge distillation with detailed settings referred in Section~\ref{appendix:lora}. As shown in Table~\ref{table: human_text}, human corpus consists of text in medical and legal domain matching the distillation knowledge base in token size. The baseline detectors refers to the same setting in Section~\ref{section: analysis_distill} where  we employ Mistral-7B-v0.2-Instruct as $\cal{M^*}$ to represent the "without distillation" scenario where the detector failed in gaining specialized knowledge. In the ablation study on detector components, details of LLM architectures are listed in Table~\ref{table:base_llm}. All detectors have the same settings in domain knowledge distillation.

\begin{table}[t]
    \centering
    \resizebox{\linewidth}{!}{
    \begin{tabular}{l p{0.45\linewidth} p{0.25\linewidth}}
        \toprule
        Knowledge Base & Seed Knowledge Composition &  Tokens Size\\
        \midrule
        Human (med) &  iCliniq~\cite{icliniq}, MedAlpaca~\cite{medalpaca}, MedRAG~\cite{medrag} & 9.5M\\
        Human (law) & Indian-Legal-Text~\cite{IndianLegalTextABS}, LegalLAMA~\cite{legalllama} & 19.6M \\
        Combined & Combined from above & 29.1M\\
        \bottomrule
    \end{tabular}
    }
    \caption{Domain Adaptation: Human Corpus Summary (M: Million Tokens)}
    \label{table: human_text}
\end{table}

\section{Additional Results}
This section records the additional results in the experiments.

\begin{figure*}[t]
    \centering
    \includegraphics[width=\linewidth]{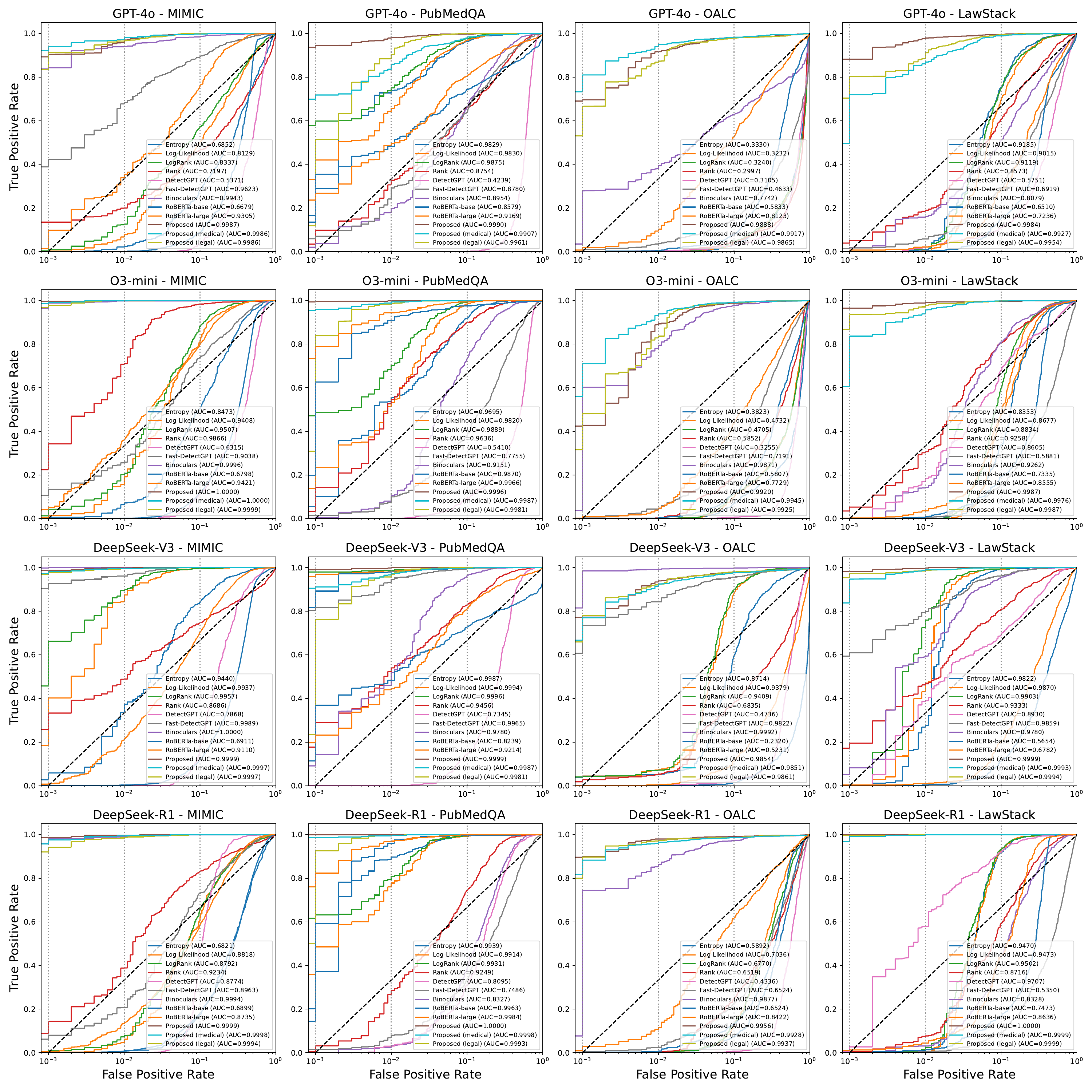}
    \caption{ROC Curves of Baseline Detectors and DivScore in the Main Experiment}
    \label{fig: auroc}
\end{figure*}

\begin{table*}[ht]
\centering
\resizebox{0.8\linewidth}{!}{%
\begin{tabular}{l|ccc|ccccc}
\toprule
\multirow{2}{*}{Methods} 
& \multicolumn{3}{c}{GPT-4o} 
& \multicolumn{3}{c}{O3-mini} \\
\cmidrule(lr){2-4} \cmidrule(lr){5-7} 
& Reddit & OpenQA & Wiki & Reddit & OpenQA & Wiki \\
\midrule
\multicolumn{7}{c}{AUROC} \\
\midrule

Fast-DetectGPT  & \underline{86.11}$_{(\downarrow1.23)}$ & 76.26$_{(\downarrow4.83)}$ & 96.29$_{(\downarrow1.92)}$ & 79.41$_{(\uparrow2.83)}$ & 61.24$_{(\uparrow4.76)}$ & 89.31$_{(\downarrow4.17)}$ \\
Binoculars  & \textbf{87.34} & 81.09 & \textbf{98.21} & 76.58 & 56.48 & 93.48    \\
\midrule
DivScore (med) & 79.22$_{(\downarrow8.12)}$ & \textbf{97.10}$_{(\uparrow16.01)}$ & 94.84$_{(\downarrow3.37)}$ & \underline{88.90}$_{(\uparrow12.32)}$ & \underline{97.27}$_{(\uparrow40.79)}$ & \underline{98.86}$_{(\uparrow5.38)}$ \\
DivScore (law) & 80.67$_{(\downarrow6.67)}$ & \underline{96.17}$_{(\uparrow15.08)}$ & \underline{96.77}$_{(\downarrow1.44)}$ & \textbf{89.17}$_{(\uparrow12.59)}$ & \textbf{97.55}$_{(\uparrow41.07)}$ & \textbf{99.37}$_{(\uparrow5.89)}$ \\
DivScore  & 78.13$_{(\downarrow9.21)}$ & 95.96$_{(\uparrow14.87)}$ & 90.72$_{(\downarrow7.49)}$ & 88.02$_{(\uparrow11.44)}$ & 97.19$_{(\uparrow40.71)}$ & 97.70$_{(\uparrow4.22)}$ \\
\midrule
\multicolumn{7}{c}{TPR @ 0.1\% FPR} \\
\midrule

Fast-DetectGPT  & \textbf{14.93}$_{(\uparrow14.33)}$ & 1.50$_{(\uparrow1.2)}$ & 21.62$_{(\uparrow0)}$ & 6.61$_{(\uparrow6.31)}$ & 0.10$_{(\uparrow0)}$ & 4.39$_{(\uparrow1.06)}$ \\
Binoculars     & 0.60 & 0.30 & 21.62 & 0.30 & 0.10 & 3.33 \\
\midrule
DivScore (med)  & 6.31$_{(\uparrow5.71)}$ & \textbf{17.65}$_{(\uparrow17.35)}$ & \underline{28.27}$_{(\uparrow6.65)}$ & 12.53$_{(\uparrow12.23)}$ & \underline{11.52}$_{(\uparrow11.42)}$ & \underline{54.75}$_{(\uparrow51.42)}$ \\
DivScore (law)  & \underline{12.73}$_{(\uparrow12.13)}$ & 5.22$_{(\uparrow4.92)}$ & \textbf{57.84}$_{(\uparrow36.22)}$ & \textbf{26.85}$_{(\uparrow26.55)}$ & 1.90$_{(\uparrow1.8)}$ & \textbf{81.83}$_{(\uparrow78.5)}$ \\
DivScore & 4.41$_{(\uparrow3.81)}$ & \underline{16.25}$_{(\uparrow15.95)}$ & 13.90$_{(\downarrow7.72)}$ & \underline{16.43}$_{(\uparrow16.13)}$ & \textbf{15.43}$_{(\uparrow15.33)}$ & 43.94 $_{(\uparrow40.61)}$ \\
\bottomrule
\end{tabular}%
}
\caption{General Domain LLM-generated Text Detection Performance, AUROC (\%) \& TPR (at 0.1 \% FPR) of selected methods on HC3-reddit, HC3-OpenQA, HC3-Wiki dataset with GPT-4o and O3-mini as text source LLMs. Methods are compared against the Binoculars benchmark, with arrows indicating performance differences (↑/↓) relative to Binoculars.}
\vspace{0.8em}
\label{table: general_domain}
\end{table*}

\subsection{Main Experiment}
\label{section: appendix_mainexp}
We present the main experiment's additional visualization results in this appendix section. Figure~\ref{fig: auroc} visualizes the AUROC curves of all baseline detectors and the proposed detector in the main experiments. The x-axis (false positive rate) uses a $\log_{10}$ scale to better illustrate detection performance at low false positive rates. This visualization complements the main experiment's metric (TPR at 0.1\% FPR) by providing a more comprehensive view of low-FPR performance.

As shown in Figure~\ref{fig: auroc}, the proposed detector enhanced with knowledge distillation in medical, legal domain and their combination achieves the best overall AUROC score in all test cases. Notably, under strict FPR thresholds (below $10^{-3}$ to $10^{-1}$), the performance gaps between our detector and baseline methods become particularly significant. This indicates DivScore's ability to detect LLM-generated profession texts with minimal false positive harm.

\subsection{Detection Performance on General Domain Texts}
The main focus of this study is to detect LLM-generated content in specialized domains. In this appendix section, we further evaluate the proposed detector's performance in general domain LLM-generated text detection which is more well studied by current research. We choose the HC3 dataset~\cite{hc3} for LLM-generated text detection. Specifically, we evaluate three sub-datasets: Reddit posts, Open-QA pairs and Wikipedia content in HC3, covering general LLM-generated text detection scenarios in social media, daily question answering and general encyclopedia content facing the public.

We employs the same settings in the main experiments, randomly selecting 1000 human texts from each datasets and 1000 LLM texts generated with GPT-4o and O3-mini. The prompt is in the QA format shown in Table \ref{table: qa_prompt}. We then benchmark the proposed detector with two state-of-the-art general domain detectors: Fast-DetectGPT~\cite{fastdetectgpt} and Binoculars~\cite{Binoculars}. As shown in Table~\ref{table: general_domain}, the proposed detector demonstrates strong performance across general domain datasets. Notably, DivScore (law) achieve substantial AUROC and TPR improvements over Binoculars on detecting O3-mini generated texts, with the averaged AUROC (\%) gains reaching 19.79 and TPR (\%) gains reaching 35.63. For GPT-4o paradigm, the proposed method, while showing slightly lower AUROC scores in Reddit and Wiki dataset, still achieves an overall higher recall at low false positive rates.

These results confirm that DivScore, though tailored for specialized domains, generalizes well beyond specialized domains. It remains highly competitive and often superior to SOTA methods in general domain LLM-generated text detection tasks.

\subsection{Case Study}
This section present four representative case studies, drawn from key experimental datasets — MIMIC, PubMedQA, OALC, and LawStack — with GPT-4o serving as the source LLM.
These cases were chosen for their high level of detection difficulty. Notably, for all four cases, state-of-the-art detectors, including Binoculars~\cite{Binoculars} and Fast-DetectGPT~\cite{fastdetectgpt}, failed to identify the generated content.

We start from the MIMIC dataset, containing de-identified hospital discharge summaries. The hospital discharge summary typically includes information about the patient's admission, diagnosis, treatment, response to treatment, and discharge plan. It is a crucial medical document that extends beyond hospital use, playing a pivotal role in commercial sectors such as healthcare insurance by ensuring precise patient care documentation and facilitating efficient claims processing. Table~\ref{table: mimic_case} demonstrates the selected MIMIC detection case with GPT-4o as the text source LLM.

In the MIMIC case, GPT-4o generated a continuation of the discharge report using the background information from the first third of the text as the prompt. Enhanced by LLM medical domain knowledge distilled from DeepSeek-R1, the proposed detector created a significant scoring gap between the LLM-generated and authentic human text ($0.103241 \to 0.275714$), indicating confident zero-shot classification. The main contribution to such gap is the Cross-Entropy component, where base and enhanced LLM show great disagreement on LLM-generated text ($4.218750$) while holding a similar probability estimation on human text ($2.734375$). This case is expected to be correctly classified with detector threshold set by below 0.1\% FPR (i.e., fewer than 1 false-positive case in detection).

Table~\ref{table: oalc_case} presents a case from the OALC dataset, with formal legislative and judicial documents from Australia government and courts. Legal documents such as statutes and regulations are characterized by their precise language, formal structure, and strict logical consistency. In this OALC case, the LLM-generated continuation closely mirrors the formal tone and structure of the original court document, with similar structure ans linguistic characteristic, leading to detection failures in Binoculars and Fast-DetectGPT.  The proposed detector, leveraging domain-specific legal knowledge distilled from DeepSeek R1, distinguishes the LLM-generated text from the authentic human-written document with a clear scoring gap ($0.163043 \to 0.084449$). The decrease entropy, again plays the key role in the detection ($0.820312 \to 0.443359$). With the detector threshold set at less than 0.1\% FPR, both the human and LLM-generated texts are confidently and correctly classified, demonstrating the detector’s robustness in handling highly specialized legal content.

Table~\ref{table: pubmed_case} shows the case selected from PubMedQA. PubMedQA contains paired questions and answers extracted from biomedical research papers in the PubMed database. Texts with similar forms of biomedical research QAs play a crucial role in  public health promotion, doctor training and evidence-based medical practice. In the PubMedQA case, enhanced by LLM medical domain knowledge distilled from DeepSeek-R1, the proposed detector created a notable scoring gap between the LLM-generated and authentic human text ($0.272876 \to 0.116497$). Such gap is attributed to a significant smaller entropy estimation in LLM-generated text ($0.535156$), compared to human text ($1.304688$). The entropy gap could be the result of a successful knowledge distillation on LLM's medical knowledge. Enhanced with LLM medical knowledge, the scoring LLM is expected to show lower entropy in LLM-generated answers, as the model tends to produce more predictable and consistent patterns compared to the broader diversity found in human writing. This finding highlights the role of entropy component in the proposed detection framework.

Table~\ref{table: lawstack_case} shows a case from the LawStack dataset, which contains legal question-answer pairs covering a wide range of jurisdictions and legal topics. Such QA pairs are commonly used in legal consulting, compliance training, and public legal education. In this case, the LLM-generated answer receives nearly one-third of detection score compared with the human-written answer, indicating a confident classification from DivScore. Such gap, as shown in Table~\ref{table: lawstack_case}, is a collective result of both scoring components. With entropy and cross-entropy both shifting as expected, the DivScore detector could successfully classify both LLM and human texts at the most restrictive threshold (FPR below 0.1\%).

\begin{table*}[ht]
    \begin{tabular}{c}
        \begin{tcolorbox}[
            colback=gray!10,
            colframe=black,
            width=\linewidth,
            boxrule=0.8pt,
            arc=1mm,
            auto outer arc,
            left=2mm,
            right=2mm,
            top=2mm,
            bottom=2mm,
            breakable
        ]
        \footnotesize
        \setlength{\parindent}{0pt}
        \setlength{\parskip}{3pt}
      
        \textbf{MIMIC: Hospital Discharge Summary}
        \vspace{4pt}\noindent\hrule height 0.4pt\vspace{4pt}
      
        \textbf{System Prompt:} Continue this hospital discharge summary.

  \textbf{User Prompt:}\\
    Name: \_\_\_ Unit No: \_\_\_ Admission Date: \_\_\_ Discharge Date: \_\_\_ Date of Birth: \_\_\_ Sex: F 
    Service: OBSTETRICS/GYNECOLOGY Allergies: Patient recorded as having No Known Allergies to Drugs Attending: \_\_\_. Chief Complaint: none Major Surgical or Invasive Procedure: none History of Present Illness: \_\_\_ yo G1P0 at 9 weeks gestation with T1 DM, sent by Dr. \_\_\_ for diabetic control. She was on Lantus and Humalog prior to pregnancy and on NPH and Humalog during pregnancy. Reports fingersticks have been average 250's. She feels well and is without complaints. Past Medical History: PRENATAL COURSE (1)LMP \_\_\_ (2)No initial prenatal yet PAST OBSTETRIC HISTORY G1 PAST GYNECOLOGIC HISTORY - no abnl paps or STIs PAST MEDICAL HISTORY - T1DM PAST SURGICAL HISTORY - ureteral reimplantation at age \_\_\_ Social History: \_\_\_ Family History: non-contributory Physical Exam: VITALS: T 96.8, HR 96, BP 128/70, RR 26, O2 100\%RA GENERAL: NAD HEART: RRR LUNGS: CTAB ABDOMEN: soft, NT EXTREMITIES: NT/NE PELVIC US: live IUP confirmed 
    
    Pertinent Results: \_\_\_ WBC-8.3 RBC-4.26 Hgb-13.0 Hct-37.0 MCV-87 Plt-314 \_\_\_ WBC-9.5 RBC-4.30 Hgb-13.2 Hct-36.5 MCV-85 Plt-314 \_\_\_ Neuts-61.8 \_\_\_ Monos-4.8 Eos-2.0 Baso-0.3 \_\_\_
    
    \vspace{4pt}\noindent\hrule height 0.4pt\vspace{4pt}
    
    \textbf{Human Text:}\\
     BLOOD HBsAg-NEGATIVE, HIV Ab-NEGATIVE \_\_\_ URINE pH-7 Hrs-24 Volu-1500 Creat-86 TotProt-<6 \_\_\_ URINE 24Creat-1290 RUBEOLA ANTIBODY, IgG (Final \_\_\_ POSITIVE BY EIA) RAPID PLASMA REAGIN TEST (Final \_\_\_ NONREACTIVE) Rubella IgG/IgM Antibody (Final \_\_\_ POSITIVE by Latex Agglutination) EARLY OB ULTRASOUND IMPRESSION: Single live intrauterine gestation. Size equals dates. 
    \\ 
    Brief Hospital Course: \_\_\_ G1 with poorly controlled T1DM admitted at 9wks for glycemic control. Ms \_\_\_ had fingersticks ranging from 151-236 on arrival to the emergency room. She had no complaints and electrolytes were normal. Her hemoglobin A1C was 10.0\%. She was admitted to the antepartum floor and \_\_\_ consulted and continued to follow her closely throughout this admission. Initially, it was unclear whether her recent elevated fingersticks were due to insulin omission or increased requirements. Nutrition was consulted. Her regimen was titrated to achieve optimal glycemic control. She was continued on NPH in the am as well as an NPH sliding scale at night. She had a humalog sliding scale with meals. Her regimen was increased and by hospital day \#4 her fingersticks were improved. She was discharged home and will have close outpatient management with \_\_\_ and the Diabetes in Pregnancy clinic. She was instructed to call with fingersticks greater than 200 for two hours. During this admission, Ms \_\_\_ had an early OB ultrasound which showed a live IUP, size equal to dates. Prenatal labs were done and were unremarkable. She was seen by social services due to her uncertain psychiatric history. Please see OMR note for details. 
    \\
    Medications on Admission: prenatal vitamins NPH 36 units qam and 20 units qhs Humalog per sliding scale with meals Discharge Medications: 1. Insulin NPH Human Recomb Subcutaneous 2. Please follow the insulin doses and sliding scale as detailed on the attached sheet. 3. Humalog Subcutaneous 4. B Complex-Vitamin C-Folic Acid 1 mg Capsule Sig: One (1) Cap PO DAILY Discharge Disposition: Home Discharge Diagnosis: Type 1 Diabetic admitted for glycemic control Discharge Condition: stable Discharge Instructions: Please call your ob doctor please call if you experience vaginal bleeding, leakage of fluid or any other concerns. Please continue current insulin regimen. Followup Instructions: \_\_\_
    
    \textbf{DivScore $ D_{\cal{M},\cal{M^{*}}}(x)$:} $0.275714$, \textbf{Entropy $H_{\cal{M^{*}}}(x)$:} $ 0.753906$, \textbf{Cross-Entropy $CE_{\cal{M},\cal{M^{*}}}(x)$:} $2.734375$
    \vspace{4pt}\noindent\hrule height 0.4pt\vspace{4pt}
    
    \textbf{LLM Text:}\\
    Patient discharged with the following instructions and management plan: Discharge Medications: 1. Insulin Therapy: Continue current regimen of NPH and Humalog as managed during the hospital stay. Patient provided with updated dosage chart and instructed to monitor blood glucose levels at home. 2. Prenatal Vitamins: One tablet daily. 3. Folic Acid: 0.4 mg daily. 
    \\
    Discharge Instructions: 1. Glycemic Control: Regular blood sugar monitoring at home. Aim for fasting blood glucose levels of 70-95 mg/dL and postprandial levels <140 mg/dL. Contact healthcare provider if experiencing episodes of hypoglycemia or persistent hyperglycemia. 2. Dietary Recommendations: Continue diabetic-friendly diet. Referral to a registered dietitian for nutritional counseling has been made. 3. Prenatal Care: Schedule an initial prenatal appointment with Dr. \_\_\_. Regular prenatal visits are important for monitoring maternal and fetal health. 4. Activity Restrictions: No specific restrictions. Engage in normal physical activity as tolerated. 5. Warning Signs: Seek immediate medical attention if experiencing symptoms such as abdominal pain, vaginal bleeding, decreased fetal movement, or signs of hyperglycemia/hypoglycemia. 
    \\
    Follow-Up: 1. Primary Care Provider: Appointment scheduled for follow-up on insulin management. 2. Obstetrician: Appointment scheduled within 1 week post-discharge for prenatal evaluation and further obstetric care planning. Prognosis: The patient is stable on discharge with well-controlled Type 1 Diabetes Mellitus in pregnancy. Regular follow-up and adherence to treatment plan will support a healthy pregnancy outcome. Conclusion: The patient was discharged in good condition, with detailed instructions provided for managing her diabetes and prenatal care. She exhibited understanding and agreement with the discharge plan. Please contact Dr. \_\_\_ or the OB/GYN clinic at \_\_\_ for any further questions or concerns regarding this discharge plan.
    
    \textbf{DivScore $ D_{\cal{M},\cal{M^{*}}}(x)$:} $0.103241$, \textbf{Entropy $H_{\cal{M^{*}}}(x)$:} $0.435547$, \textbf{Cross-Entropy $CE_{\cal{M},\cal{M^{*}}}(x)$:} $4.218750$
    \vspace{4pt}\noindent\hrule height 0.4pt\vspace{4pt}
    \textbf{Detector Threshold}@0.1\%FPR: $0.206166$
        \end{tcolorbox}
    \end{tabular}
    \caption{MIMIC Case Study: Comparison of human and LLM generated hospital discharge summary with detection scores}
    \label{table: mimic_case}
\end{table*}

\begin{table*}[ht]
\centering
\tiny
    \begin{tabular}{c}
        \begin{tcolorbox}[
            colback=gray!10,
            colframe=black,
            width=\linewidth,
            boxrule=0.8pt,
            arc=1mm,
            auto outer arc,
            left=2mm,
            right=2mm,
            top=2mm,
            bottom=2mm,
            breakable
        ]
        \footnotesize
        \setlength{\parindent}{0pt}
        \setlength{\parskip}{3pt}
        
        \textbf{OALC: Open Australia Legal Corpus}
        
        \vspace{4pt}\noindent\hrule height 0.4pt\vspace{4pt}
        
        \textbf{System prompt:} Continue write this legal text.

        \textbf{User prompt:}\\
        FEDERAL COURT OF AUSTRALIA
        Van der Velde v Ng [2008] FCA 480   TERRY VAN DER VELDE AND DAVID STIMPSON AND RIBY PTY LTD (ACN 081 153 219) IN LIQUIDATION AS TRUSTEE FOR THE JADE TRUST v KYM HON YOKE NG, THE OFFICIAL TRUSTEE AS TRUSTEE OF THE PROPERTY OF ROSS HASTINGS (A BANKRUPT) AND RAMS MORTGAGE CORPORATION LIMITED ABN 48 065 912 932 QUD 405 OF 2007  DOWSETT J  4 FEBRUARY 2008  BRISBANE IN THE FEDERAL COURT OF AUSTRALIA  QUEENSLAND DISTRICT REGISTRY       QUD 405 OF 2007   BETWEEN:  TERRY VAN DER VELDE AND DAVID STIMPSON            First Applicant            RIBY PTY LTD (ACN 081 153 219) IN LIQUIDATION AS TRUSTEE FOR THE JADE TRUST            Second Applicant  AND:      KYM HON YOKE NG            First Respondent            THE OFFICIAL TRUSTEE
        \vspace{4pt}\noindent\hrule height 0.4pt\vspace{4pt}
        \textbf{Human Text:}\\
        AS TRUSTEE OF THE PROPERTY OF ROSS HASTINGS (A BANKRUPT)            Second Respondent            RAMS MORTGAGE CORPORATION LIMITED ABN 48 065 912 932            Third Respondent    JUDGE:  DOWSETT J  DATE:   4 FEBRUARY 2008  PLACE:  BRISBANE   REASONS FOR JUDGMENT     
        
        1 In both Fraser v Deputy Commissioner of Taxation \& Official Trustee (1996) 69 FCR 99 and Scott v Bagshaw (1999) 95 FCR 37, the relevant applicant was seeking to enforce a provable debt although, in the latter case, that may not have appeared from the statement of claim. In those circumstances, s 58(3) of the Bankruptcy Act 1966 (Cth) clearly applied.     
        
        2 That is not the present case, as I understand it. In those circumstances, I am of the view that leave is not necessary. However, should in the course of the trial it emerge otherwise, we can revisit the question. I certify that the preceding two (2) numbered paragraphs are a true copy of the Reasons for Judgment herein of the Honourable Justice Dowsett.    
        
        Associate: Dated: 14 April 2008      
        Solicitor for the First Applicant:    Mills Oakley  
        Counsel for the Second Applicant:     The Second Applicant did not appear  
        Counsel for the First Respondent:     The First Respondent did not appear  
        Solicitor for the Second Respondent:  Rodgers, Barnes \& Green  
        Counsel for the Third Respondent:     Mr D S Piggott  
        Solicitor for the Third Respondent:   Mallesons Stephen Jacques  
        Date of Hearing:                      4 February 2008  
        Date of Judgment:                     4 February 2008

        \textbf{DivScore $ D_{\cal{M},\cal{M^{*}}}(x)$:} $0.163043$, \textbf{Entropy $H_{\cal{M^{*}}}(x)$:} $0.820312$, \textbf{Cross-Entropy $CE_{\cal{M},\cal{M^{*}}}(x)$:} $5.03125$
        \vspace{4pt}\noindent\hrule height 0.4pt\vspace{4pt}
        \textbf{LLM Text:} \\
        AS TRUSTEE OF THE PROPERTY OF ROSS HASTINGS (A BANKRUPT)            Second Respondent            RAMS MORTGAGE CORPORATION LIMITED ABN 48 065 912 932            Third Respondent   JUDGE:  DOWSETT J  DATE OF ORDER:  4 FEBRUARY 2008   WHERE MADE:  BRISBANE  REASONS FOR JUDGMENT  
        
        Introduction  
        1. This matter involves a complex set of legal proceedings concerning multiple parties and a series of claims arising from financial transactions, property interests, and trust management. The applicants, Terry Van der Velde and David Stimpson, along with Riby Pty Ltd acting as trustee for the Jade Trust, seek remedies against the respondents, who include Kym Hon Yoke Ng, the Official Trustee managing the bankruptcy estate of Ross Hastings, and RAMS Mortgage Corporation Limited.  
        
        Background  
        2. The facts of the case date back to transactions undertaken in 2005, where the applicants allege financial mismanagement and breaches of fiduciary duty by the first respondent, leading to significant financial detriments and potential loss of property interests administered under the Jade Trust.  
        
        3. The involvement of the Official Trustee arises from the concurrent bankruptcy proceedings associated with Ross Hastings, necessitating the management of his financial affairs, while the third respondent, RAMS Mortgage Corporation Limited, is implicated due to its interests in mortgage securities tied to the disputed property.  
        
        Legal Issues  
        4. The key legal issues revolve around the fiduciary duties allegedly breached by the first respondent, the implications of these breaches on the trust’s financial stability, the proper scope of actions available by the liquidated Riby Pty Ltd against the respondents, and the determination of priority among the creditors, including RAMS Mortgage Corporation Limited.  
        
        5. Additional questions concern the jurisdiction of this Court in managing disputes related to bankruptcy and asset recovery, with particular attention to cross-claims presented by the second respondent regarding the equitable distribution of assets.  
        
        Proceedings and Orders  
        6. The Court will undertake a detailed examination of the evidence presented, including financial statements, transaction records, and testimonial accounts provided by all parties. This examination will allow for an informed judgement on breaches of duty and mismanagement claims.  
        
        7. Interim orders will be considered to safeguard the property interests of the applicants pending final resolution, including potential injunctions against foreclosure actions by mortgage holders.  
        
        Conclusion  
        8. The outcomes of this case will clarify the obligations of trustees and managers of financial trusts and provide guidance for creditor priorities within bankruptcy contexts, contributing to the broader understanding of fiduciary and financial stewardship in corporate and trust law.  
        * * *  
        I certify that the preceding text forms the official reasons for judgment handed down by the Honourable Justice Dowsett, on the 4th day of February, 2008.  
        [Signature of Justice Dowsett]

        \textbf{DivScore $ D_{\cal{M},\cal{M^{*}}}(x)$:} $0.084449$, \textbf{Entropy $H_{\cal{M^{*}}}(x)$:} $0.443359$, \textbf{Cross-Entropy $CE_{\cal{M},\cal{M^{*}}}(x)$:} $5.25000$
        \vspace{4pt}\noindent\hrule height 0.4pt \vspace{4pt}
        \textbf{Detector Threshold}@0.1\%FPR: $0.118119$
        \end{tcolorbox}
    \end{tabular}
    \caption{OALC Case Study: Comparison of Human and LLM-generated Legal document with detection scores}
    \label{table: oalc_case}
\end{table*}

\begin{table*}[htbp]
\tiny
    \begin{tabular}{c}
        \begin{tcolorbox}[
            colback=gray!10,
            colframe=black,
            width=\linewidth,
            boxrule=0.5pt,
            arc=1mm,
            auto outer arc,
            left=2mm,
            right=2mm,
            top=2mm,
            bottom=2mm,
            breakable
        ]
        \footnotesize
        \setlength{\parindent}{0pt}
        \setlength{\parskip}{3pt}
        
        \textbf{PubMedQA: Answer to Biomedical Research Questions}
        \vspace{4pt}\noindent\hrule height 0.4pt\vspace{4pt}
        
        \textbf{System Prompt:} Answer this question truthfully.

        \textbf{User Prompt:} Is adjustment for reporting heterogeneity necessary in sleep disorders?
        \vspace{4pt}\noindent\hrule height 0.4pt\vspace{4pt}
        
        \textbf{Human Text:} \\
        Sleep disorders are common in the general adult population of Japan. Correction for reporting heterogeneity using anchoring vignettes is not a necessary tool for proper management of sleep and energy related problems among Japanese adults. Older age, gender differences in communicating sleep-related problems, the presence of multiple morbidities, and regular exercise should be the focus of policies and clinical practice to improve sleep and energy management in Japan.
        
        \textbf{Detection Score $ D_{\cal{M},\cal{M^{*}}}(x)$:} $0.272876$, \textbf{Entropy $H_{\cal{M^{*}}}(x)$:} $1.304688$, \textbf{Cross-Entropy $CE_{\cal{M},\cal{M^{*}}}(x)$:} $4.78125$
        \vspace{4pt}\noindent\hrule height 0.4pt\vspace{4pt}
        
        \textbf{LLM Text:} \\
        Adjustment for reporting heterogeneity can be necessary in sleep disorders research to ensure accurate and reliable results. Reporting heterogeneity refers to variations in how information is documented or reported, which can stem from differences in assessment tools, study design, participant demographics, cultural differences, and subjective interpretations of symptoms. In sleep disorders, symptoms and their severity can be perceived differently by individuals due to personal, cultural, or situational factors. Without adjusting for these differences, there can be inconsistencies in data interpretation, potentially leading to skewed or misleading findings. Therefore, employing methods to adjust for reporting heterogeneity can enhance the validity and comparability of research outcomes across different studies or populations.

        \textbf{DivScore $ D_{\cal{M},\cal{M^{*}}}(x)$:} $0.116497$, \textbf{Entropy $H_{\cal{M^{*}}}(x)$:} $0.535156$, \textbf{Cross-Entropy $CE_{\cal{M},\cal{M^{*}}}(x)$:} $4.59375$
        \vspace{4pt}\noindent\hrule height 0.4pt\vspace{4pt}
        \textbf{Detector Threshold}@0.1\%FPR: $0.130639$
        \end{tcolorbox}
    \end{tabular}
    \caption{PubMedQA Case Study: Comparison of Human and LLM-generated answers to biomedical research questions with detection scores}
    \label{table: pubmed_case}
\end{table*}

\begin{table*}[ht]
    \centering
    \begin{tabular}{c}
        \begin{tcolorbox}[
            colback=gray!10,
            colframe=black,
            width=\linewidth,
            boxrule=0.5pt,
            arc=1mm,
            auto outer arc,
            left=2mm,
            right=2mm,
            top=2mm,
            bottom=2mm,
            breakable
        ]
        \footnotesize
        \setlength{\parindent}{0pt}
        \setlength{\parskip}{3pt}
        
        \textbf{LawStack: Answer to Legal Questions}
        \vspace{4pt}\noindent\hrule height 0.4pt\vspace{4pt}
        
        \textbf{System Prompt:} \\
        Answer this question truthfully:
        
        \textbf{User Prompt:} \\
        I have read the question and answer here: Are parking tickets enforcable But I would like to know that if a person got a parking ticket in Pennsylvania for a street sweeper violation and didn't know until they got a registered letter from the city. In that letter, they state that a person can plead not guilty (and send in a bond amount) or guilty. Which leads me to think that it's a criminal offense and not a civil one. If it's a criminal offense, wouldn't the burden of proof be on the city to prove that a particular person (and not just the owner) did the parking? Or is it really sufficient to serve the person who they think is the owner? Can they really issue a warrant for arrest if the owner might not have ever set foot in that state before? If personal service isn't made (as outlined in the above question) then is a parking ticket in PA really enforceable? What if the owner of the car doesn't live in Pennsylvania and instead lives in Texas? Can they be extradited to PA from TX or placed under arrest somehow?
        \vspace{4pt}\noindent\hrule height 0.4pt\vspace{4pt}
        
        \textbf{Human Text:} \\
        The procedure depends on the local rules. For example, Philadelphia. In Philadelphia, parking offences are civil, not criminal - other governments may be different. Irrespective of if it is civil or criminal, the burden is on the city. All that is different is the burden: for civil it is the preponderance of the evidence (or balance of probabilities), for criminal it is beyond reasonable doubt. Under Philadelphia law, anyone driving with the owner\'s permission is an agent of the owner so serving them is serving the owner. There is an exemption that car rental companies can avail themselves of. The can take whatever enforcement action the law allows - this may include issuing an arrest warrant. Serving by post or affixing to the vehicle is legally service. If there is a valid warrant, they can be extradited from anywhere subject to relevant extradition law - most extradition laws require the crime to be "serious" - parking violations probably aren't.

        \textbf{DivScore $ D_{\cal{M},\cal{M^{*}}}(x)$:} $0.203187$, \textbf{Entropy $H_{\cal{M^{*}}}(x)$:} $0.796875$, \textbf{Cross-Entropy $CE_{\cal{M},\cal{M^{*}}}(x)$:} $3.921875$
        \vspace{4pt}\noindent\hrule height 0.4pt\vspace{4pt}
        
        \textbf{LLM Text:} \\
        In Pennsylvania, as in many other jurisdictions, parking violations, including those for street sweeper offenses, are generally considered civil infractions rather than criminal offenses. This means that the violation is typically against the vehicle and not directly against the individual who parked it. As such, the burden of proof is generally on the city to show that the vehicle was parked in violation of the law, rather than proving who specifically parked the car.  The registered letter you received offering options to plead not guilty or guilty and pay a bond amount is a common procedural step in handling such civil infractions. While parking tickets are civil matters, failure to address them can lead to additional consequences, such as fines, vehicle registration holds, or even a bench warrant for failing to respond to the citation, which can create legal complications.  Regarding jurisdiction, if the owner of the vehicle resides out of state, such as in Texas, Pennsylvania cannot extradite the individual solely for an unpaid parking ticket. Extradition typically applies to criminal offenses. However, unresolved parking tickets can lead to complications like increased fines or affecting the owner's ability to register their vehicle, depending on interstate compacts or agreements.  If you believe that the ticket was issued in error or you have compelling reasons to challenge it, you may wish to contest the ticket through the legal process outlined in the letter. It can also be helpful to seek legal advice to understand the specific options and implications based on the circumstances of the case.
        
        \textbf{DivScore $ D_{\cal{M},\cal{M^{*}}}(x)$:} $0.078887$, \textbf{Entropy $H_{\cal{M^{*}}}(x)$:} $0.404297$, \textbf{Cross-Entropy $CE_{\cal{M},\cal{M^{*}}}(x)$:} $5.125000$
        \vspace{4pt}\noindent\hrule height 0.4pt\vspace{4pt}
        \textbf{Detector Threshold}@0.1\%FPR: $0.099856$
        \end{tcolorbox}
    \end{tabular}
    \caption{LawStack Case Study: Comparison of Human and LLM-generated answers to legal issue with detection scores}
    \label{table: lawstack_case}
\end{table*}

\end{document}